\documentclass{article}

\usepackage[preprint]{neurips_2026}

\usepackage[utf8]{inputenc} % allow utf-8 input
\usepackage[T1]{fontenc}    % use 8-bit T1 fonts
\usepackage{hyperref}       % hyperlinks
\usepackage{url}            % simple URL typesetting
\usepackage{booktabs}       % professional-quality tables
\usepackage{amsfonts}       % blackboard math symbols
\usepackage{nicefrac}       % compact symbols for 1/2, etc.
\usepackage{microtype}      % microtypography
\usepackage{xcolor}         % colors
\usepackage{graphicx}
\usepackage{listings}
\usepackage{xcolor}
\definecolor{codeblue}{RGB}{245,248,255}   % light blue background
\usepackage{tcolorbox}
\tcbuselibrary{breakable}
\usepackage{fvextra}
\usepackage{wrapfig}
\usepackage{placeins}
\usepackage{amsmath}
\usepackage{makecell}
\usepackage{subcaption}
\title{CuTeGen: An LLM-Based Agentic Framework for Generation and Optimization of High-Performance GPU Kernels using CuTe}

\author{%
  Tara Saba \\
  Department of Computer Science\\
  University of Toronto\\
  \texttt{tara.saba@mail.utoronto.ca} \\
  \AND
  Zhiyang Chen \\
  Department of Computer Science\\
  University of Toronto\\
  \texttt{zhiychen@cs.toronto.edu} \\
  \AND
  Jikai Jason Li \\
  Department of Computer Science\\
  University of Toronto\\
  \texttt{jasonjikai.li@mail.utoronto.ca} \\
  \AND
   Anne Ouyang \\
  Standard Kernel\\
    \texttt{anne@standardkernel.com} \\
  \AND
   Xujie Si \\
   Department of Computer Science\\
  University of Toronto\\
    \texttt{six@cs.toronto.edu} \\
   \AND
   Fan Long \\
   Department of Computer Science\\
  University of Toronto\\
   \texttt{fanl@cs.toronto.edu} \\
}

\begin{document}

\maketitle

\begin{abstract}
High-performance GPU kernels are critical to modern machine learning systems, yet developing them remains a manual, expert-driven process. Recent work has explored using LLMs to automate kernel generation, but generated kernels still fall short of carefully tuned references on standardized benchmarks. We present CuTeGen, an agentic GPU kernel synthesis framework that treats kernel development as a structured generate–test–refine workflow over the CuTe abstraction layer. Two design choices distinguish CuTeGen from prior work: targeting CuTe rather than raw CUDA, which exposes performance-critical structures such as tiling and data movement while remaining stable enough for iterative refinement, and a delayed profiling schedule that withholds low-level performance feedback until the kernel's high-level structure has stabilized. On the 209 tasks of KernelBench Level-1 and Level-2, CuTeGen achieves an average speedup of 1.71$\times$
over PyTorch and outperforms the prior agentic baseline CudaForge (0.89$\times$) at comparable per-task generation cost. Code available at https://github.com/taratt/cutegen.git 
\end{abstract}

%%%%%%%%%%%%%%%%%%%%%%%%%%%%%%%%%%%%%%%%%%%%%%%%%%%%%%%%%%%%
\section{Introduction}
The continued scaling of deep learning models—especially large language models (LLMs)—has made high-performance GPU computing a central bottleneck for modern AI systems. Across workloads in language, vision, and scientific computing, end-to-end throughput is often dominated by a small set of compute-intensive primitives. While GPU hardware capabilities have advanced rapidly, realizing these gains in practice depends critically on low-level kernel implementations that effectively exploit massive parallelism, the memory hierarchy, and specialized instructions. As a result, system performance is frequently determined not by algorithms alone, but by how close kernel code can approach the hardware’s theoretical limits.

LLM-driven coding agents have reshaped many aspects of software development, yet their impact on high-performance GPU kernels remains limited. For many performance-critical operators, e.g., GEMM and FlashAttention \cite{dao2022flashattention}, state-of-the-art implementations are still largely hand-engineered by experts. Recent work has explored using LLMs to generate GPU kernels automatically, but the performance gap between generated kernels and carefully tuned expert implementations is still substantial, especially for kernels that rely on hardware-specific features and tightly coupled optimization decisions~\cite{lange2025towards, liao2025kernelevolve, ouyang2025kernelbench}.

A core reason is that efficient GPU kernels are extremely sensitive to low-level design choices. High performance typically requires jointly selecting (i) hardware instructions and tensor-core usage (e.g., MMA/WMMA variants), (ii) work decomposition and tiling across thread blocks and warps, (iii) explicit management of data movement and caching (e.g., shared-memory layouts and bank-conflict avoidance), and (iv) software pipelining to overlap memory transfers with computation (e.g., staged buffering and latency hiding). These choices are highly interdependent: changing a tile shape affects shared-memory layout, which affects instruction selection and scheduling constraints, which in turn affects occupancy and achievable throughput. This creates a large, coupled search space in which single-shot code generation is unlikely to find a good combination and naive iterative edits often break correctness or regress performance.

\noindent \textbf{CuTeGen:} This paper presents CuTeGen, an agentic GPU kernel synthesis system that iteratively generates, evaluates, debugs, and refines kernels via a structured execution-feedback loop. Instead of treating kernel synthesis as a one-time generation problem, CuTeGen treats it as an autonomous optimization process: candidate kernels are compiled, tested against reference outputs, and timed; compilation diagnostics, runtime errors, correctness discrepancies, and performance signals are then fed back to guide subsequent iterations. CuTeGen separates debugging from optimization and emphasizes incremental edits instead of full rewrites, enabling progressive correction and performance improvement without requiring access to expert-written kernels or manual intervention from experienced GPU developers.

A key design choice in CuTeGen is to generate kernels in CuTe~\cite{cute}, a C++/CUDA template abstraction layer that exposes performance-critical structures such as tiling, layout, and data movement, while providing enough scaffolding to make generation and iterative refinement more stable than in raw CUDA. By operating in CuTe, CuTeGen can more naturally leverage tensor-core-friendly tiling strategies and incorporate memory-pipelining patterns that are central to high-performance GEMM kernels, thereby biasing the search space toward more efficient implementations. Moreover, unlike higher-level domain-specific languages such as Triton~\cite{tillet2019triton}, CuTe preserves the low-level control needed for fine-grained hardware-specific optimization, including direct integration with CUDA features such as inline PTX.

Finally, to guide performance tuning, CuTeGen integrates hardware profiling using NVIDIA Nsight Compute~\cite{nsight}. Importantly, CuTeGen employs a \emph{delayed profiling integration} technique and does not expose profiling metrics from the outset. For structurally complex kernels such as matrix multiplication, we delay profiling-driven feedback until the code has reached a reasonable baseline through higher-level structural optimization. We find that introducing profiling too early often encourages myopic parameter tuning, such as adjusting tile sizes, before the kernel’s overall structure is sound, thereby increasing the risk of premature convergence to poor local optima. Once the kernel has stabilized, curated profiling summaries are introduced to support targeted refinement.

\noindent \textbf{Results:} We evaluate CuTeGen on the level-1 and level-2 benchmark set of kernels from KernelBench and compare against CudaForge~\cite{zhang2025cudaforge}, the prior state-of-the-art agentic GPU kernel generation framework. Across the 209 evaluated kernels, CuTeGen generated kernels that outperformed the PyTorch reference for 76 tasks, while CudaForge did so for only 45 tasks.

\noindent \textbf{Contributions:} This paper makes the following contributions:
\begin{itemize}
\item \textbf{CuTeGen:} We present CuTeGen, a novel iterative GPU kernel synthesis framework that generates kernels in CuTe. By operating in CuTe, CuTeGen guides LLMs toward a search space that is richer in efficient kernels, while retaining the low-level control needed for further kernel optimization.

\item \textbf{Delayed Profiling Integration:} We propose a delayed profiling integration technique that allows CuTeGen to incorporate profiling metrics for performance tuning without prematurely driving kernel generation toward poor local optima.
\item \textbf{Empirical Evaluation:} We perform a comprehensive evaluation on all 209 KernelBench Level-1 and Level-2 tasks and show that CuTeGen substantially outperforms the prior agentic baseline CudaForge. Our ablation studies further demonstrate the importance of operating in CuTe and delaying profiling-driven feedback.
\end{itemize}

\section{Background}

High-performance GPU kernels are typically developed using vendor libraries,
template libraries, domain-specific programming systems, and compiler frameworks.
Vendor-provided libraries 
such as cuBLAS~\cite{cublas} and cuDNN~\cite{chetlur2014cudnn} 
offer highly optimized implementations of common operations but 
are closed-source and limited to predefined kernels. 
Template-based libraries such as
CUTLASS~\cite{cutlass} offer
reusable CUDA templates for building custom kernels, while still requiring
substantial manual engineering effort. Higher-level systems such as
Triton~\cite{tillet2019triton} and ThunderKittens~\cite{spector2024thunderkittens}
make kernel programming more accessible by abstracting memory layout and
execution structure, but developers have 
less control over low-level optimization choices.
Compiler-based systems such as
\texttt{torch.compile}~\cite{paszke2019pytorch} further automate graph- and
operator-level transformations, but generally only optimize 
using fixed optimization patterns. These systems reduce the 
burden of GPU kernel
development, yet efficient custom kernels still often depend on 
expert-designed
implementations and optimization strategies.

CuTeGen targets the automatic generation and optimization of custom GPU kernels from PyTorch reference implementations. To support this goal, CuTeGen uses CuTe~\cite{cute}, the tensor abstraction layer used by CUTLASS, as its kernel representation. CuTe provides structured abstractions for layouts, tiling, tensors, and data movement while preserving the low-level control required for hardware-aware optimization. This combination gives the model a stable representation for iterative refinement without sacrificing access to performance-critical GPU features.

\section{Related Work}

\paragraph{Early Studies of LLM Kernel Generation Capabilities.}
Early works~\cite{godoy2023evaluation, valero2023comparing} take the first steps toward empirically evaluating the ability of LLMs to generate high-performance computing
kernels. 
BabelTower~\cite{wen2022babeltower} and
CodeRosetta~\cite{tehrani2024coderosetta} instead focus on translating
sequential or high-level programs into parallel implementations. These works
establish the feasibility of LLM-assisted parallel programming, but primarily
study generation capabilities or program translation. CuTeGen targets a more
specific and challenging problem. 

\paragraph{Training-based and search-based kernel generation.}
Another line of work improves kernel generation through model training or
structured search. AscendKernelGen~\cite{cao2026ascendkernelgen},
AutoTriton~\cite{li2025autotriton}, and CUDA-L1~\cite{li2025cuda} adapt models
with domain-specific data, supervised fine-tuning, or reinforcement learning to
improve hardware-specific code generation. K-Search~\cite{cao2026k} and
AutoComp~\cite{hong2025autocomp} instead explore optimization spaces at
inference time, the former using a co-evolving world model and the latter
LLM-driven search with hardware feedback. CuTeGen differs from these systems on
two axes. Where the training-based approaches rely on model adaptation and the
search-based approaches on parallel space exploration, CuTeGen iteratively
refines a single evolving kernel via execution feedback, with no retraining and
no parallel search. The two groups also target different abstraction layers---Triton DSL for AutoTriton and Python/PyTorch code for AutoComp---whereas
CuTeGen operates at the CuTe C++/CUDA template layer, which exposes tiling and
layout while preserving low-level hardware control.

\paragraph{Hardware-feedback and agentic kernel optimization.}
The closest related systems use execution or profiling feedback to iteratively
improve GPU kernels. Some works operate primarily on CUDA kernels:
CUDA-LLM~\cite{chen2025cuda} uses compilation, correctness, and runtime feedback
to refine CUDA implementations; CUDAForge~\cite{zhang2025cudaforge} adds a
Coder-Judge workflow with Nsight Compute feedback; and Astra~\cite{wei2025astra}
coordinates specialized agents to optimize existing CUDA kernels. CuTeGen differs
from these systems by using CuTe as the synthesis representation, exposing layout
and tiling structure to the model instead of refining raw CUDA code directly.
Other works target other abstractions or broader search spaces.
TritonForge~\cite{li2025tritonforge} applies profiling-guided optimization to
Triton kernels, while KernelEvolve~\cite{liao2025kernelevolve} explores evolving
candidate implementations across multiple accelerator programming abstractions,
including Triton and CuTe. CuTeGen differs from these systems by centering the
entire workflow on a single CuTe kernel and improving it through localized patch
edits, rather than maintaining a broad candidate search over many alternative
implementations.

\section{Method}
\label{sec:method}

CuteGen operates as an agentic refinement loop in which kernels generated 
by a large language model are compiled, executed, validated for correctness, 
and iteratively improved. The workflow is organized into two main phases:
\textbf{Correctness Assurance Stage} shown in Figure~\ref{fig:correctness}
, and \textbf{Optimization Stage} shown in Figure~\ref{fig:optimization}.

Starting from an initial task specification and reference implementation, 
the model produces a CuTe-based kernel. This kernel is compiled, executed, and 
validated against the reference implementation using randomized inputs. 
Compilation failures, runtime errors, and output mismatches are recorded 
and used to guide subsequent refinement steps. 
When correctness issues arise, diagnostic information is incorporated into 
structured debugging prompts that enable the model to analyze failures and generate 
targeted code patches. Once a kernel produces correct outputs, the system transitions 
to the optimization phase, where performance feedback obtained through hardware profiling, 
together with a domain-specific optimization guide that we provide to the model, 
is used to guide further improvements. Through this iterative feedback loop, 
CuTeGen progressively refines both functional correctness and execution efficiency. 

\paragraph{Task Formulation and Objective.}
CuTeGen takes as input a PyTorch specification of the target computation, 
typically written as a \texttt{torch.nn.Module} whose \texttt{forward} method 
defines the intended semantics. The specification may additionally provide helper 
code for constructing representative input tensors, mainly to specify expected tensor shapes, 
data types, and invocation patterns. Based on this specification, CuTeGen synthesizes 
a functionally equivalent implementation in which performance-critical PyTorch operators 
are replaced by custom GPU kernels. The output of CuTeGen is therefore not 
just a generated kernel, but an optimized implementation that can be invoked under the 
same high-level interface as the original PyTorch code. From the user's perspective, 
adopting CuTeGen only requires providing a correct reference implementation together 
with representative inputs; CuTeGen then handles kernel generation and iterative 
optimization toward improved execution performance. This task formulation also aligns 
with standardized evaluation setups used in benchmarks such as 
KernelBench~\cite{ouyang2025kernelbench}.

In our framework, we explicitly guide the model to generate kernels using 
the \textbf{CuTe} abstraction layer. CuTe provides structured tensor and 
layout abstractions that expose performance-critical design choices such as 
tiling strategies, memory layouts, and execution organization while preserving 
low-level control over GPU execution. Compared to raw CUDA, these 
abstractions make kernel generation more tractable for language models, 
while remaining lower-level than DSLs such as Triton or ThunderKittens.
 This combination provides a stable representation for iterative refinement 
 while retaining the flexibility needed for hardware-aware optimization.

% \begin{table}[t]
% \centering
% \small
% \caption{Comparison of GPU programming abstractions across optimization capabilities and control over hardware execution. 
% Entries indicate whether a given abstraction explicitly exposes a capability to the developer or code generator. 
% This comparison highlights the design trade-off between structured guidance and low-level control that motivates our use of CuTe.\\} 
% \label{tab:abstraction_comparison}

% \begin{tabular}{lccccccc}
% \toprule
% \textbf{Abstraction} &
% \textbf{Abstraction} &
% \textbf{Explicit} &
% \textbf{Shared} &
% \textbf{Instruction} &
% \textbf{Tensor/Layout} &
% \textbf{Hardware} \\
%  &
% \textbf{Level} &
% \textbf{Tiling} &
% \textbf{Memory} &
% \textbf{Access} &
% \textbf{Abstractions} &
% \textbf{Control} \\
% \midrule

% Python (PyTorch)
% & Very High
% & No
% & No
% & No
% & No
% & Low \\

% Triton
% & High
% & Partial
% & Abstracted
% & Limited
% & Yes
% & Medium \\

% \textbf{CuTe}
% & Structured Low
% & Yes
% & Yes
% & Yes
% & Yes
% & High \\

% CUDA
% & Low
% & Yes
% & Yes
% & Yes
% & No
% & Very High \\

% PTX
% & Assembly
% & Yes
% & Yes
% & Yes
% & No
% & Maximum \\

% \bottomrule
% \end{tabular}
% \end{table}
\subsection{Correctness Assurance Stage}

CuTeGen follows an iterative generate--test--refine workflow in which kernel 
correctness is established before performance optimization is attempted. The 
correctness assurance stage coordinates a sequence of LLM interactions and 
program executions designed to progressively identify and repair errors in 
generated kernels.

The process begins with an \emph{initial synthesis prompt} 
% (depicted in Figure \ref{fig:init_prompt}), 
(detailed in Appendix~\ref{app:initial_prompt})
which describes 
the target task and provides a representative CuTe kernel example to anchor the 
expected structure and coding style. Using this prompt, the LLM produces an 
initial candidate kernel implementation.

\begin{figure}[t]
  \centering
  \includegraphics[width=\linewidth]{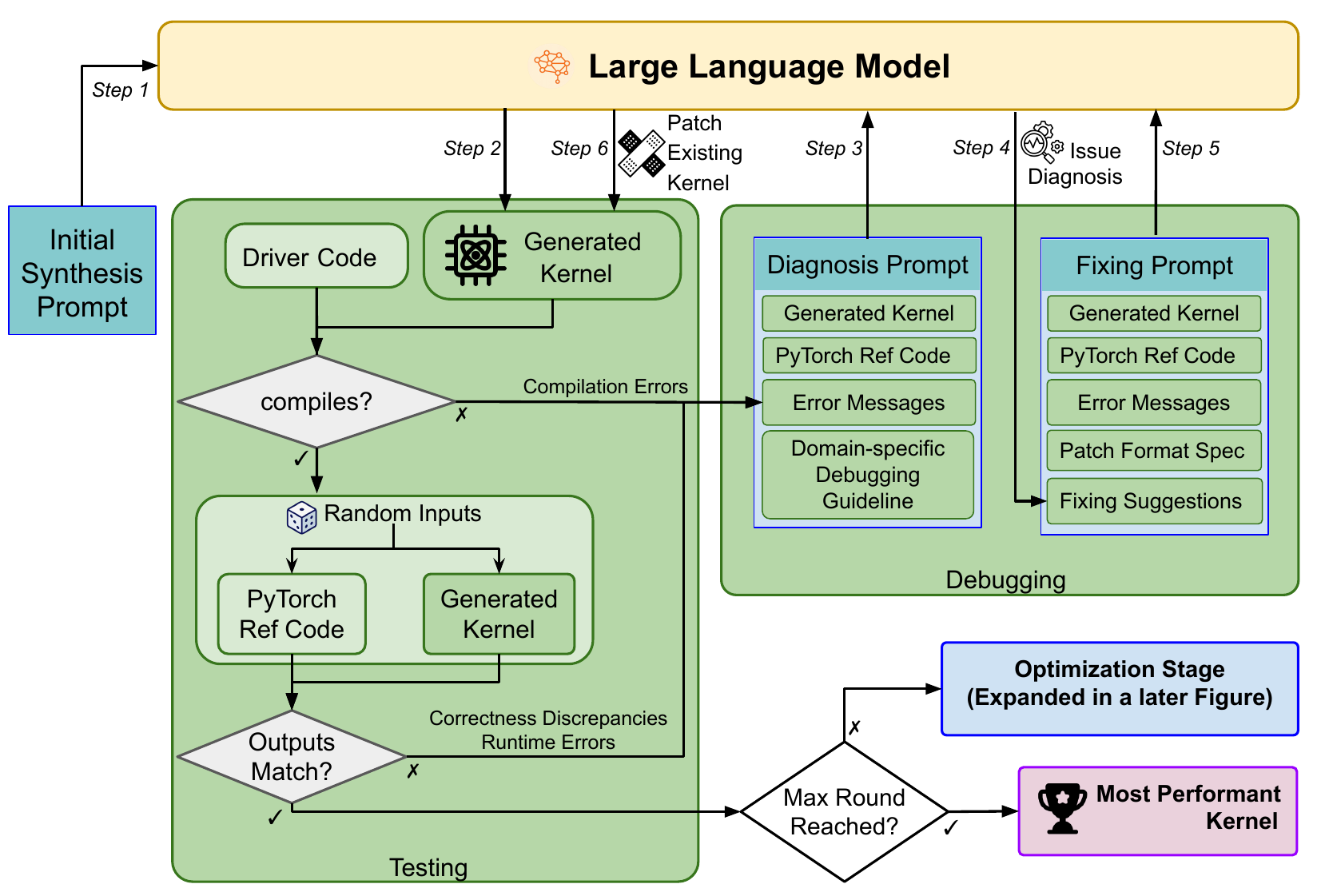}
\caption{
Detailed view of the correctness assurance component. Generated kernels are compiled and validated against reference outputs, and structured patches are applied iteratively until correctness is achieved or the generation budget is exhausted.
}
  \label{fig:correctness}
\end{figure}

The generated kernel is then evaluated within the \emph{correctness assurance 
stage} (shown in Figure~\ref{fig:correctness}). In the \emph{testing phase}, the system repeatedly 
compiles, executes, and validates the implementation against reference PyTorch 
outputs using randomized inputs. Compilation failures produce compiler 
diagnostics, while successful executions are checked for output equivalence with 
the reference implementation. Any compilation errors, runtime failures, or 
correctness discrepancies detected during testing trigger the
\emph{debugging process}.

When errors are detected, CuTeGen invokes a structured debugging interaction 
organized as a two-stage process: (\emph{i}) diagnosis and 
(\emph{ii}) repair. This separation reflects our design goal 
of encouraging explicit reasoning about failures before modifying the kernel 
implementation.
\paragraph{Diagnosis Stage.}
In the diagnosis stage, the model receives diagnostic information—including 
compilation errors, runtime exceptions, and correctness discrepancies—together 
with our structured debugging guidelines, and is prompted to generate 
\emph{diagnostic suggestions} explaining the likely causes of the observed 
failures. We developed these guidelines based on both domain knowledge of CuTe 
kernel development and empirical observations of common failure patterns in 
LLM-generated kernels. Accordingly, they encode both CuTe-specific 
considerations and recurrent LLM pitfalls, providing targeted guidance for 
interpreting compiler feedback, runtime behavior, and correctness mismatches. 
The full debugging guideline used in our system is provided in 
Appendix~\ref{app:debugging_guidelines}.

\paragraph{Repair Stage.}
Following diagnosis, the model generates \emph{structured patch edits} that 
repair the implementation. Rather than regenerating the entire kernel, the model 
proposes localized modifications in a constrained patch format (e.g., line-level 
insertions, deletions, or replacements), which are then applied directly to the 
existing code. This staged design encourages the model to first reason explicitly 
about the source of failure before proposing edits, leading to more targeted and 
coherent fixes. Constraining the repair process to localized patches further 
preserves valid portions of the existing kernel, avoids unnecessary large-scale 
rewrites, and helps retain performance-relevant design choices while resolving 
correctness issues.

Once a kernel passes compilation and correctness validation, the optimization 
stage is invoked to improve performance. Subsequent optimization steps may 
again introduce correctness issues, in which case the same execution, validation, 
and debugging loop is re-entered. This alternating process ensures that 
performance improvements are built on functionally correct implementations. 

\subsection{Optimization Stage}
\label{sec:opt}
\begin{figure}[t]
  \centering
  \includegraphics[width= 0.8\linewidth]{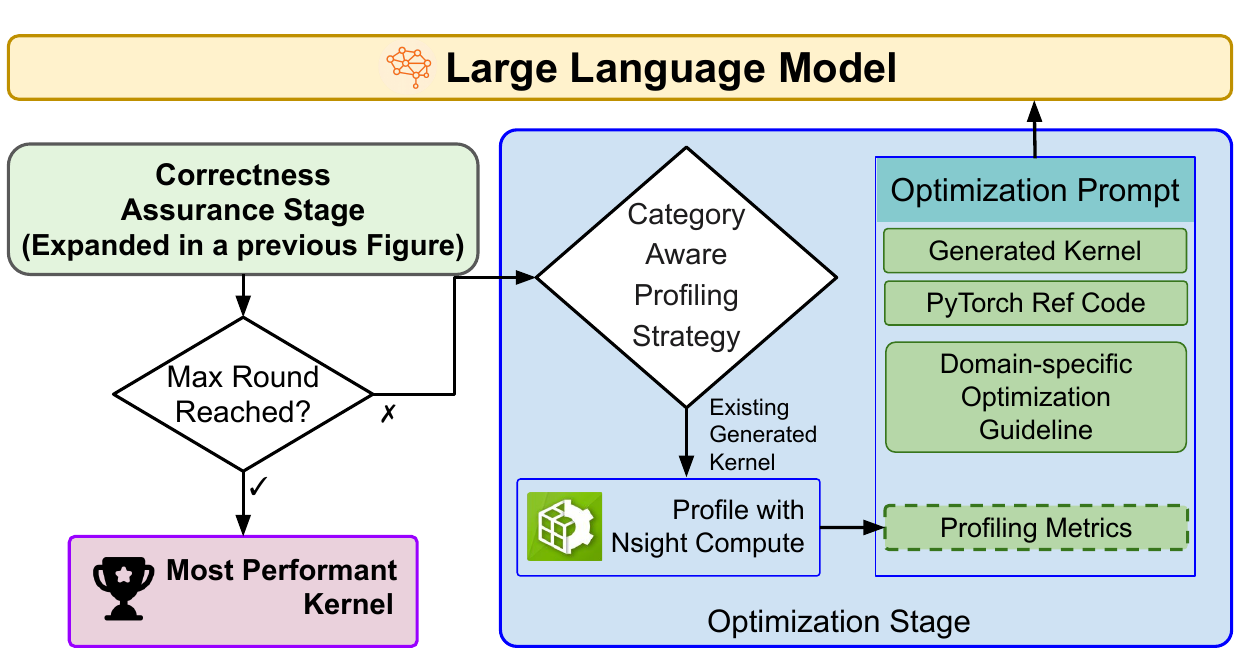}
\caption{
Detailed view of the optimization component. Profiling feedback is introduced according to a workload-dependent strategy to guide performance-aware refinement.
}
  \label{fig:optimization}
\end{figure}

Once a generated kernel successfully compiles and produces correct outputs, CuTeGen 
transitions from correctness assurance to performance optimization (shown in Figure \ref{fig:optimization}). We conceptualize 
this stage as comprising two complementary components: prompt-guided structural 
optimization and profiling-driven refinement.

\paragraph{Structural Optimization.}
In the structural optimization phase, CuTeGen prompts the underlying LLM to 
generate performance-aware kernel transformations while preserving correctness. 
The prompt specifies a set of \emph{task-specific optimization guidelines} based on 
GPU kernel design principles and empirical observations of LLM optimization behavior. 
CuTeGen uses different optimization guidelines for different kernel categories. The model is instructed to infer 
the kernel category, reason about dimensional relationships and relevant GPU optimization 
strategies, and then apply the corresponding optimization principles. For example, matrix 
multiplication kernels are guided toward tiling strategies, memory hierarchy utilization, 
and data layout transformations, whereas activation-function kernels emphasize vectorization, 
memory access patterns, and lightweight computation fusion. The full optimization prompts 
used in our system are provided in Appendix~\ref{app:optimization_prompt}.

Importantly, we encourage the model to introduce optimizations incrementally—typically 
one transformation at a time—rather than attempting multiple simultaneous changes. 
Based on our observations, this strategy improves stability, reduces correctness 
regressions, and requires fewer debugging iterations compared to more aggressive 
multi-step modifications. This incremental refinement behavior is facilitated by the 
structured starting point provided by the CuTe representation, which exposes tiling and 
execution structure from the outset; a direct comparison with a raw CUDA baseline is 
provided in Appendix~\ref{app:cute_vs_cuda}. In practice, this structured and category-aware guidance 
enables the model to prioritize meaningful algorithmic improvements over superficial 
code changes. A detailed GEMM optimization case study is provided in Appendix \ref{sec:case_study_square_gemm}.

\paragraph{Profiling-Driven Refinement.}
In the profiling-driven refinement phase, CuTeGen incorporates hardware profiling 
signals obtained using NVIDIA Nsight Compute to inform later optimization stages. 
Rather than exposing raw profiler 
outputs, we provide the model with a structured summary of performance indicators, 
including kernel launch configuration parameters (e.g., block and grid sizes, 
register usage, and shared-memory allocation), execution-time metrics such as 
kernel duration and throughput estimates, resource utilization measures including 
achieved occupancy, and selected profiler diagnostics 
highlighting potential bottlenecks. Presenting these signals in a concise,
structured form enables the model to reason about performance trade-offs without 
being overwhelmed by low-level detail.

A key design choice in our system, especially for primitive operations, is the \emph{delayed introduction of profiling feedback}. We do not expose profiling signals uniformly across all workloads. For structurally complex kernels such as matrix multiplication, profiling information is intentionally withheld during early optimization iterations and introduced only after the model has explored higher-level structural transformations. We found that introducing profiling feedback too early biases the model toward premature parameter tuning (e.g., tile sizes or shared-memory configurations), often leading to suboptimal local optima before more impactful algorithmic improvements are implemented. By delaying profiling, we encourage the model to first establish a strong structural design and only then refine low-level parameters.

In contrast, for workloads whose optimization opportunities are less structurally complex and more parameter-driven, such as activation functions, profiling feedback is introduced earlier to facilitate efficient tuning.
% This workload-dependent scheduling of profiling signals allows CuTeGen to balance global algorithmic exploration with targeted performance refinement.

%Together, these components enable CuTeGen to systematically combine high-level 
%algorithmic optimization with informed low-level tuning, resulting in stable and 
%progressively improving kernel synthesis within the agentic refinement loop.
\section{Experiments}
\label{sec:exp}
This section evaluates CuTeGen. Our evaluation is organized around two research questions:
\begin{itemize}
\item \textbf{RQ1 (Effectiveness):} Does CuTeGen generate kernels that are both correct and faster than the PyTorch reference, and how does it compare against CudaForge, a state-of-the-art agentic kernel-generation system?

\item \textbf{RQ2 (Design choices):} How much of CuTeGen's performance is attributable to its core design decisions — generating in CuTe rather than raw CUDA, and the delayed integration of profiling feedback?
\end{itemize}

%\subsection{Experimental Setup}

\paragraph{Benchmarks.}
KernelBench~\cite{ouyang2025kernelbench} is a benchmark suite designed to evaluate whether 
LLMs can generate correct and high-performance GPU kernels from PyTorch 
reference implementations. 
We evaluate CuTeGen on all 209 tasks from KernelBench Level-1 and Level-2.

Level-1 consists of single primitive operations — matrix multiplication, convolution, activation functions, normalization and reduction layers, and loss functions. Because PyTorch dispatches these directly to hand-tuned vendor libraries such as cuBLAS and cuDNN, the PyTorch reference is already a heavily optimized kernel, and achieving speedup over it is very challenging.

Level-2 combines 3--6 primitives into a single workload (e.g., convolution followed by bias addition and an activation). PyTorch eager launches each primitive as a separate kernel, so the most effective optimization is often \emph{operator fusion} — combining the sequence into one kernel to eliminate the overhead of multiple kernel launches and intermediate memory traffic.

\paragraph{Comparison Baseline: CudaForge.}
We surveyed several open-source systems from prior and concurrent work on agentic GPU kernel generation, and select CudaForge~\cite{zhang2025cudaforge} as our baseline. Among the systems we considered, CudaForge is the only one that is both publicly available and can be driven by the same underlying language model (GPT-5) we use for CuTeGen, allowing a fair comparison in which the model and hardware are held constant and only the agent design varies. The remaining candidates were unsuitable for one of three reasons: (i) the system requires training and is bound to its own fine-tuned model, precluding a head-to-head GPT-5 comparison; (ii) the system operates outside the kernel itself — for instance, modifying the surrounding PyTorch program rather than producing a standalone GPU kernel — and is therefore not directly comparable to CuTeGen's task formulation; or (iii) we were unable to reproduce the system in our environment. We run CudaForge with its default 10 round setting.

\paragraph{Hardware and Software Environment.}
All experiments are conducted on a workstation equipped with a single NVIDIA GeForce RTX 4090 GPU (24 GB memory) with CUDA 13.0 support. Generated kernels are compiled using PyTorch 2.8.0 through its CUDA extension interface together with CUTLASS v4.3.0 (commit acb4593), which provides the CuTe abstraction layer used in our system. Kernel generation and refinement are performed using GPT-5. All methods are evaluated under identical hardware and software conditions.

\paragraph{Evaluation Protocol.}
We report performance as the speedup of each generated kernel over the reference PyTorch implementation, computed as the ratio of reference runtime to generated kernel runtime. All baselines are evaluated against PyTorch \emph{eager} execution, with no compiler-driven fusion or graph-level optimization, so that measured speedups reflect kernel design and in-kernel fusion rather than graph-level rewrites. To obtain accurate timings, we warm up the GPU before each measurement to amortize initialization and caching effects. 

\paragraph{Manual Correctness Audit.}
Tolerance-based correctness checks alone are insufficient for evaluating LLM-generated kernels. When pushed to optimize aggressively, language models frequently take shortcuts that compromise numerical accuracy — most commonly, enabling mixed-precision tensor core paths by lowering operand precision from FP32 to FP16/BF16 — and such kernels may pass the benchmark's tolerance check while silently producing degraded outputs that would be unacceptable in deployment. To guard against this, we manually inspect every generated kernel, with particular scrutiny on those reporting a speedup over the PyTorch reference, for both CuTeGen and CudaForge. Kernels whose speedup is attributable to lowering precision below the reference, or otherwise altering the underlying computation, are excluded from the reported speedup numbers.

\subsection{Main Results}

Table~\ref{tab:cutegen_vs_cudaforge} summarizes the performance results of kernels generated by CuTeGen and CudaForge. The column ``Speedup'' denotes the average speedup achieved across all 209 tasks relative to the PyTorch reference implementation. The
column ``Low Precision'' reports the percentage of tasks for which the reported best kernel reduces the computation precision to achieve higher performance. The column ``Correct'' denotes the percentage
of tasks for which the tool generated at least one correct kernel that passed both automated testing and manual inspection. ``Fast1'' reports the percentage of tasks for which the generated kernel outperforms the PyTorch reference implementation. The column ``Speedup On Fast1'' reports the average speedup obtained when using the generated kernels only for tasks where they outperform PyTorch, while retaining the PyTorch implementation for the remaining tasks. Column ``Best'' denotes the percentage of tasks where the corresponding tool generated an optimized kernel that is best among all three versions (i.e., PyTorch, CuTeGen, and CudaForge). The column ``Cost (USD)'' denotes the average token cost of running the tool per task.

\paragraph{Overall Performance.}
The results in Table~\ref{tab:cutegen_vs_cudaforge} show that
CuTeGen can generate highly efficient kernels and consistently outperforms CudaForge. CuTeGen achieves an average speedup of 1.71$\times$
on its generated kernels, whereas CudaForge achieves only 0.89$\times$.
Out of the 209 cases, CuTeGen generates the best-performing kernel in 34\% of the tasks while CudaForge does so in only 11\% of the cases. The user can use the CuTeGen generated kernels to obtain a much higher speedup (2.16$\times$ versus 1.45$\times$). The cost of CuTeGen per case is only slightly higher than CudaForge but still very acceptable at 1.69 dollar per task when running on GPT-5. 

One interesting observation is that CudaForge generated lower-precision kernels in a substantial fraction of tasks --24\% of the cases on which it reports speedup-- often by enabling FP16/BF16 computations while retaining FP32 accumulation. In contrast, CuTeGen does not exhibited this behavior, partly because our prompts across the debugging and optimization stages explicitly instruct the model to preserve the precision in computation. We found that many mixed-precision kernels can still pass the tolerance-based random testing harness used in KernelBench, despite deviating from the numerical behavior of the reference implementation. While such changes may improve benchmark speedups, they can silently reduce numerical reliability and lead to unacceptable behavior when deployed as part of larger end-to-end systems.

\paragraph{Level-1 Performance.}
Level-1 is the most demanding setting: each task is a standalone primitive and the PyTorch reference dispatches directly to cuBLAS or cuDNN, which are at or near hardware peak. Beating this reference requires precise control over tiling, shared-memory use, and instruction scheduling, and is the regime in which most LLM-generated kernels fail. On these tasks, CuTeGen consistently achieves higher speedup than CudaForge, and outperforms it most clearly on the categories where the vendor library is most aggressive. We attribute this gap to two design choices: operating in CuTe rather than raw CUDA exposes tiling and layout decisions to the model in a structured form, and delaying profiling integration lets the model establish a sound algorithmic skeleton before tuning low-level parameters.

\paragraph{Level-2 Performance.}
On Level-2, CuTeGen still outperforms CudaForge overall, but CudaForge can generate better kernels than CuTeGen often.
Level-2 workloads are dominated by a single optimization lever — operator fusion across a sequence of primitives — and once both systems successfully fuse, the remaining headroom from low-level tuning is small relative to the gain from collapsing kernel launches. The advantages CuTeGen accrues on Level-1 from its structured representation matter less here, because the dominant decision (whether and how to fuse) is well within reach of either approach. 

\paragraph{Variance of Speedup Results.}
Across both levels, the per-task speedup distribution exhibits high variance and is not fully captured by category averages. For example, for most matrix multiplication shapes, neither system outperforms PyTorch, as highly optimized vendor libraries such as cuBLAS leave little room for improvement. Consequently, average gains within this category are driven primarily by a small number of structured workloads where general-purpose vendor kernels are far from optimal. On diagonal matrix multiplication, for instance, CuTeGen achieves a 16.35$\times$ speedup while CudaForge achieves 10.43$\times$. These results illustrate that average performance should be interpreted together with the underlying task-level distribution rather than as uniformly representative across workloads.

\begin{table}[t]
\centering
\caption{
Comparison of CuTeGen and CudaForge on KernelBench Level-1 and Level-2 workloads.
We report average speedup, precision lowering rate (\%), correctness (\%), Fast1 (\%), speedup on Fast1 tasks, percentage of best-performing kernels (\%), and average cost per kernel (USD).
}
\label{tab:cutegen_vs_cudaforge}

\footnotesize

\begin{tabular*}{\linewidth}{@{\extracolsep{\fill}}lccccccc}
\toprule
\textbf{Tool}
& \textbf{Speedup} $\uparrow$
& \textbf{Low Precision} $\downarrow$
& \textbf{Correct} $\uparrow$
& \textbf{Fast1} $\uparrow$
& \textbf{Speedup On Fast1} $\uparrow$
& \textbf{Best} $\uparrow$
& \textbf{Cost (USD)} $\downarrow$ \\
\midrule
CudaForge 
& 0.89$\times$& 24\% & 63\%
& 22\% & 1.45$\times$ & 11\%
& 1.41 \\

\midrule
\textbf{CuTeGen}
& 1.71$\times$ & 0\% & 87\%
& 36\% & 2.16$\times$ & 34\%
& 1.69 \\

\bottomrule
\end{tabular*}
\end{table}

\if 0
\begin{table}[t]
\centering
\label{tab:cutegen_vs_cudaforge}

\footnotesize

\begin{tabular*}{\linewidth}{@{\extracolsep{\fill}}lccccccc}
\toprule
\textbf{Method}
& \multicolumn{3}{c}{\textbf{Level 1}}
& \multicolumn{3}{c}{\textbf{Level 2}}
& \textbf{Cost (USD)} $\downarrow$ \\
\cmidrule(lr){2-4} \cmidrule(lr){5-7}
& Speedup $\uparrow$
& Fast1 $\uparrow$
& Correct. $\uparrow$
& Speedup $\uparrow$
& Fast1 $\uparrow$
& Correct. $\uparrow$
& \\
\midrule

CudaForge 
& 0.72$\times$& 18\% & 69\%
& 1.41$\times$ & 47\% & 72\%
& 1.41 \\

\midrule
\textbf{CuTeGen}
& 1.35$\times$ & 50\% & 94\%
& 2.12$\times$ & 22\% & 76\%
& 1.69 \\

\bottomrule
\end{tabular*}
\end{table}

\fi

\subsection{Ablation Studies}
\label{ablation}

To evaluate the impact of key design decisions in CuTeGen, we conduct a study on all Level-1 KernelBench tasks using three modified variants of the system. The first variant removes all prompting related to CuTe-based kernel generation, the second disables profiling feedback entirely, and the third introduces profiling feedback from the beginning of the optimization process rather than delaying it.

Table~\ref{tab:ablation} summarizes the ablation study results in different categories of KernelBench level-1 tasks. The columns ``MatMul'', ``Activation Function'', ``Convolution'', ``Reduction/Normalization'', and ``Loss'' show the average speedup of the corresponding tool in each task category. The column ``Total'' shows the average speedup across all tasks of the corresponding tool. Our results show that the original CuTeGen outperforms all three variants. Turning off the CuTe has the largest impact, degrading the overall speedup from 1.35$\times$ to 1.12$\times$. This highlights the importance of the CuTe abstraction to guide LLM to operate with efficient tiling and pipelining patterns. Note that for the ``Reduction/Normalization'' category, profiling from the start outperforms CuTeGen. This exception is primarily driven by a single high-speedup workload that disproportionately increases the category average for that variant. Additional experimental results are provided in Appendix \ref{ab_app}.

\begin{table}[t]
\centering
\caption{
Average speedup over PyTorch on KernelBench Level-1 tasks for different CuTeGen variants, grouped by kernel category.
}
\label{tab:ablation}
\resizebox{\linewidth}{!}{
\begin{tabular}{lcccccc}
\toprule
\textbf{Method}
& \textbf{MatMul}
& \textbf{Activation Function}
& \textbf{Convolution}
& \textbf{Reduction/Normalization}
& \textbf{Loss}
& \textbf{Total} \\
\midrule

CuTeGen profiling from start
& 1.18$\times$
& 1.80$\times$
& 0.45$\times$
& \textbf{2.08$\times$}
& 4.01$\times$
& 1.34$\times$
\\
CuTeGen without profiling
& 1.47$\times$
& 1.76$\times$
& 0.57$\times$
& 1.25$\times$
& 4.04$\times$
& 1.23$\times$

\\
CuTeGen with cuda
& 1.25$\times$
& 1.79$\times$
& 0.48$\times$
& 1.05$\times$
& \textbf{4.06$\times$}
& 1.12$\times$

\\

\textbf{CuTeGen}
& \textbf{1.62$\times$}
& \textbf{1.82$\times$}
& \textbf{0.61$\times$}
& 1.42$\times$
& \textbf{4.06$\times$}
& \textbf{1.35$\times$}

\\
\bottomrule
\end{tabular}
}
\end{table}

\section{Limitations}

CuTeGen does not match expert-tuned vendor kernels  such as cuBLAS, performance-wise, on the most demanding workloads, most notably dense matrix multiplication. Owing to compute budget, our experiments use a single language model (GPT-5) on a single NVIDIA RTX 4090; the framework is model-agnostic in principle, and we expect comparable behavior on other Ada and Ampere-class GPUs. Newer architectures such as Hopper and Blackwell introduce features (e.g., TMA, wgmma) not exercised by our current prompts, and reaping additional speedup on those generations will likely require prompt updates that surface them.

\section{Conclusion}

We presented CuTeGen, an iterative GPU kernel synthesis framework that generates kernels in CuTe and refines them through a structured generate–test–profile loop. On the 209 tasks of KernelBench Level-1 and Level-2, CuTeGen outperforms the prior agentic baseline CudaForge in both average speedup and fraction of best-of-three kernels, and our ablations attribute the gap to two design decisions: targeting CuTe rather than raw CUDA, which exposes tiling and layout structure to the model, and delaying the integration of profiling feedback, which lets the model establish a sound algorithmic skeleton before tuning low-level parameters. More broadly, these results indicate that agentic systems are a promising direction for optimizing performance-critical GPU computation, and that the right choice of intermediate representation and feedback schedule materially affects what such systems can achieve.

\bibliographystyle{plain} 
\bibliography{refs}
\newpage
\appendix
\section*{Appendix}
\section{Initial Synthesis Prompt}
\label{app:initial_prompt}

\begin{wrapfigure}[23]{r}{0.30\linewidth}
\vspace{-12pt}
\centering
\includegraphics[width=\linewidth]{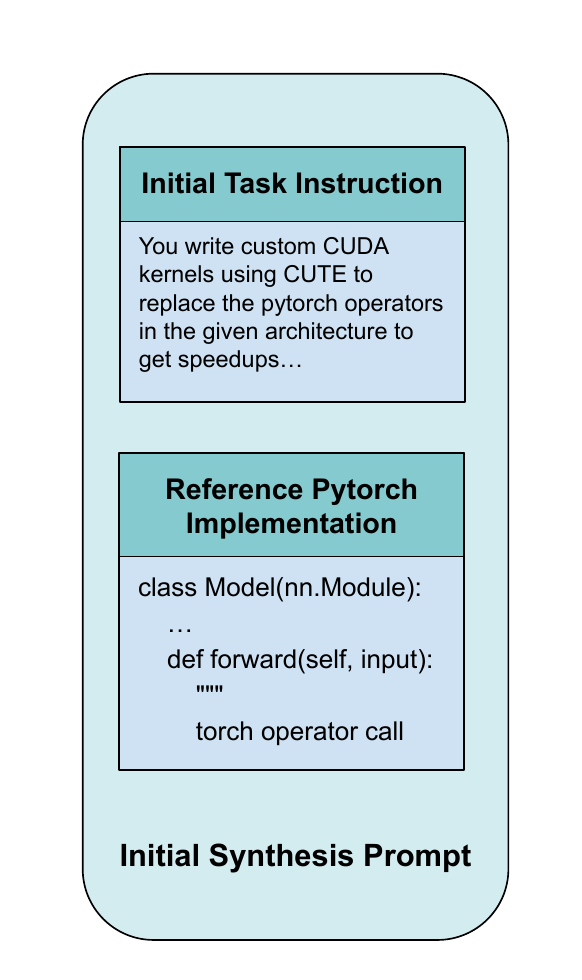}
\caption{Initial synthesis prompt used to generate the first candidate kernel.}
\label{fig:init_prompt}
\vspace{-14pt}
\end{wrapfigure}

The initial synthesis prompt (Figure~\ref{fig:init_prompt}) provides the LLM with the target PyTorch implementation together with a representative CuTe kernel example illustrating the expected integration and implementation style. The prompt instructs the model to replace the original PyTorch computation with a custom CuTe kernel while preserving the functionality of the original module. Using this information, the LLM generates an initial candidate implementation that serves as the starting point for the subsequent generate--test--refine workflow.

\section{Debugging Guidelines Used in the Correctness Assurance Component}
\label{app:debugging_guidelines}
A central element of the Correctness Assurance Component in CuTeGen is the use of a structured debugging guideline that governs how the model analyzes and repairs incorrect kernels. Rather than relying on unconstrained trial-and-error code generation, our framework provides the model with explicit debugging rules derived from common failure patterns in GPU kernel development and from recurring errors observed in LLM-generated implementations. These guidelines act as a form of operational domain knowledge that shapes the model’s reasoning during the debugging process.

The importance of this guideline arises from the nature of high-performance GPU kernels. Small implementation errors—such as mismatched tensor layouts, incorrect partitioning logic, missing synchronization, or inconsistent tile dimensions—can lead to silent correctness failures, runtime crashes, or performance regressions. In iterative synthesis settings, naïve debugging strategies often attempt to resolve such failures by simplifying the implementation (e.g., removing tiling, bypassing asynchronous pipelines, or replacing custom kernels with library calls). While these changes may restore correctness, they typically destroy the optimization structure of the kernel and prevent meaningful performance improvement. The debugging guideline therefore enforces a disciplined repair strategy that preserves the original algorithmic intent while systematically identifying and correcting the underlying source of failure.

Conceptually, the guideline serves three complementary roles in the system. First, it constrains the model’s search space during debugging by explicitly prohibiting destructive simplifications, ensuring that fixes remain consistent with the intended kernel design. Second, it provides a structured checklist of common correctness pitfalls specific to CuTe and GPU programming, enabling the model to reason about errors using domain-relevant concepts such as tensor shapes and strides, memory layouts, synchronization semantics, and hierarchical tiling. Third, it standardizes the debugging procedure itself by encouraging the model to separate diagnosis from modification, promoting targeted patch edits rather than full kernel rewrites.

The guideline encodes several categories of debugging knowledge. These include rules for preserving kernel structure and optimization intent, explanations of core CuTe abstractions such as tensors, layouts, and tiling hierarchies, and checklists for detecting common correctness issues such as indexing mismatches, synchronization errors, and memory-layout inconsistencies. It also documents common pitfalls related to asynchronous memory operations (e.g., \texttt{cp.async}), tensor-core usage, and memory allocation behavior. Together, these components provide the model with a consistent framework for reasoning about failures and generating reliable repairs during iterative refinement.

For reproducibility and transparency, we include below the exact debugging guideline used in all correctness-repair interactions in our experiments.

\begin{tcolorbox}[
  colback=white,
  colframe=black,
  boxrule=0.8pt,
  breakable,
  left=4pt,
  right=4pt,
  top=4pt,
  bottom=4pt,
  title={CuTe Coding and Debugging Guide}
]
\begin{Verbatim}[breaklines=true,fontsize=\footnotesize]
                     CuTe CODING & DEBUGGING GUIDE
===============================================================================

## NON-NEGOTIABLE RULE (READ FIRST)

**DO NOT change the intent of the code by making the CUDA/CuTe code simpler or by using less CUDA/CuTe.**
- Do **not** “fix” bugs by replacing CuTe tiling/partitioning/TiledMMA/pipeline with a naive CUDA kernel.
- Preferably do **not** remove async copies, TMA/cp.async, swizzles, tensor layouts, or CuTe abstractions just to pass correctness.
- Do **not** fall back to cuBLAS / CUTLASS device::Gemm unless the original intent *already was* to call it.
- Your job is to make the *same algorithm and same CuTe structure* correct: repair layouts, strides, partitions, synchronization, predication, and integration wiring—**without de-CuTe-ing the kernel**.
- Do **not** change the input sizes (input tensor sizes) given in the original model.

This is a correctness guide, not a “simplify to green tests” guide.

===============================================================================

## TABLE OF CONTENTS

PART 0: LLM OPERATING PROCEDURE (How to debug without changing intent)
PART 1: CORE CONCEPTS (Layouts, Shapes/Strides, Tensors, Tiles)
PART 2: CuTe KERNEL ORGANIZATION (STRUCTURE TO PRESERVE)
PART 3: CORRECTNESS DEBUGGING CHECKLIST (Silent bugs & fixes)
PART 4: ASYNC COPY & SYNC (cp.async / TMA / barriers / pipeline gotchas)
PART 5: MMA / TiledMMA PITFALLS (Atom selection, partitioning, layouts)
PART 6: MEMORY & OOM PITFALLS (Especially from redundant contiguous conversions)

===============================================================================
PART 0: LLM OPERATING PROCEDURE
===============================

When you are given incorrect CUDA/CuTe code, follow this strict workflow:

1) **Freeze intent**
   * Write down (explicitly, in the response) what the code is trying to compute:
     - inputs/outputs
     - mathematical operation (e.g., GEMM, convolution-like, reduction, attention block, stencil, etc.)
     - invariants that must remain true (tiling strategy, pipeline stages, shared-memory staging, etc.)
   * Confirm what you will NOT change (see Non-Negotiable Rule above).

2) **Freeze convolution semantic contract (REQUIRED for conv / conv-transpose / conv-like kernels)**
   * Before changing any convolution kernel, explicitly freeze:
     - the logical input layout used by the task/reference
       - e.g. `(N,C,D,H,W)` vs `(N,C,H,W,D)`
     - the logical weight layout used by the task/reference
       - e.g. `(C_out, C_in/groups, Kd, Kh, Kw)`
     - which spatial axes the kernel actually slides over
     - which axis, if any, has kernel extent 1
     - the exact output-shape formula for those axes
   * Do **NOT** silently “normalize” the kernel to standard PyTorch conventions unless the task definition itself proves those conventions.
   * If the task docstring, `get_inputs()`, reference code, and expected output shape disagree with your assumption, trust the task/reference semantics first and resolve that conflict before editing CUDA/CuTe code.
   * Required check before proposing a fix:
     - write the expected output shape from the task definition
     - write the output shape produced by the current kernel’s indexing/shape formula
     - if they differ, fix semantic/layout interpretation first, not low-level CUDA details
   * Typical failure pattern this prevents:
     - the kernel is mathematically correct for `(N,C,D,H,W)` but the task actually uses `(N,C,H,W,D)`, or vice versa
     - a `(K,K,1)` kernel is accidentally turned into a `(1,K,K)` kernel by changing axis interpretation

3) **Localize the bug without de-optimizing**
   * You may add:
     - debug prints (guarded)
     - assertions
     - extra verification kernels
     - temporary checksums / spot-check loads
     - temporary extra sync for diagnosis (then remove once fixed)
   * You may NOT replace the CuTe structure with a simpler kernel “just for debugging.”

4) **Fix in the smallest possible CuTe-native way**
   * Typical correct fixes:
     - wrong `Layout` (shape/stride mismatch)
     - wrong tile view (`local_tile`, `tile`, `partition_*`) producing permuted coordinates
     - missing predicate for edge tiles
     - missing barrier/wait for async copy stages
     - wrong smem layout for the chosen MMA atom
     - wrong accumulator type / epilogue cast timing
   * Avoid “global refactors.” Make one change, re-test.

5) **Prove the fix**
   * Provide:
     - a short statement of what was wrong
     - what changed
     - why it preserves intent
     - how it was validated (small cases + at least one edge case)
   * For CuTe GEMM-like or convolution kernels, “validated” means more than compiling:
     - confirm the logical tensor/layout/axis roles are still correct after the change
     - confirm the output-shape formula still matches the task/reference

===============================================================================
PART 1: CORE CONCEPTS
=====================

1) **Layout = (Shape, Stride) and it is the truth**
   * In CuTe, `Layout` directly defines address mapping. If output is “plausible but wrong,” assume layout/stride is wrong until proven otherwise.

2) **Tensor = pointer + Layout**
   * A CuTe `Tensor` is a view. “Correct launch” does not imply “correct indexing.”

3) **Tiling/partitioning are coordinate transforms**
   * `local_tile`, `partition_*`, etc. do not copy—they remap indices. Debug them like math.

===============================================================================
PART 2: CuTe KERNEL ORGANIZATION (STRUCTURE TO PRESERVE)
===================================================================

A correct CuTe kernel typically has these layers; **do not delete these layers**:

A) **Global tensors (gmem)**
B) **CTA tiling (block-level shapes)**
C) **Smem staging + Copy (sync or async)**
D) **Compute (MMA or per-thread math)**
E) **Epilogue / store (with correct predicates)**

For non-matmul kernels, the same structure still applies: you still have gmem views, tiles, smem staging, and compute that must be consistent.

===============================================================================
PART 3: CORRECTNESS DEBUGGING CHECKLIST
=======================================

Fix in this order:

A) **Shape/Layout/Stride sanity**
1. Confirm each tensor’s real stride (from how data is produced).
2. Confirm each CuTe view preserves the intended coordinates.
3. Spot-check a few coordinates → physical addresses → expected values.
4. Before proposing fixes, check whether the kernel assumes contiguous linear indexing (e.g., X[i], reinterpret_cast<float4*>) and whether the wrapper forces .contiguous() or changes device; if inputs may be non-contiguous, either (A) require contiguous inputs explicitly (TORCH_CHECK + contiguity gate + scalar tail) to match the reference behavior used in this harness, or (B) implement a correct strided fallback (sizes/strides + offset computation) while keeping the optimized contiguous fast path.

B) **Copy path correctness**
5. Validate copy-only loop (gmem→smem→gmem) before looking at compute.
6. If async is used, ensure required waits/barriers exist before first use.

C) **CuTe rank/projection API failures: fix the smallest semantic mismatch first**

If compilation fails with errors involving:

- `parameter "X" is not a type name`
- `Step<_1, X, ...>` / `Step<X, _1, ...>` not compiling
- `local_tile(...)` / `local_partition(...)` overload mismatch
- rank mismatch between tensor, tiler, coord, or projection
- invalid type conversion from `make_gmem_ptr(nullptr)` or dummy tensors

treat this as a CuTe API/rank-semantics bug, not a performance bug.

Debug in this order:

1. Verify tensor rank:
   - if the tensor is 2D, use 2D tilers/coords/projections
   - if the tensor is 3D, use 3D tilers/coords/projections
   - do not mix a 3D CTA tiler with a 2D tensor unless the projection is explicitly valid

2. Do not introduce new `Step<...>` forms unless the same form already exists and compiles in the current source.
   - Some CuTe builds / contexts do not accept the projected form you are trying to use.
   - Prefer simpler `local_tile(...)` / `local_partition(...)` calls first.

3. Do not build dummy CuTe tensors from null global pointers just to infer fragment shapes.
   - This often causes invalid pointer-type conversions and fragile code generation.
   - Prefer deriving layout/shape from real tensors or from explicit layouts.

4. If `local_partition(...)` fails, check logical role first:
   - Is this tensor supposed to be A, B, or C?
   - What rank and shape should the fragment have?
   - Do not patch the call blindly until you can state the intended fragment shape.

5. Apply one local fix only.
   - Do not simultaneously change CTA tiling rank, thread layout, projection syntax, and operand layout.

D) **Compute correctness**
7. Validate partition shapes match expectations (especially K mapping for MMA).
8. Verify accumulator dtype and any casting happens late enough.

E) **Edges**
9. Predication for partial tiles (M/N/K remainders) must be correct.
10. Ensure stores are masked correctly (no out-of-bounds writes).

F) **High-frequency LLM failure modes (check before anything else)**
1) Do NOT worry or check for device mismatches. Our setup is ALWAYS single-gpu.
2) Do NOT assume contiguous/stride/alignment unless you guard it (TORCH_CHECK or gated fast path + correct fallback).
3) If using float4/vec fast path: gate by alignment + stride==1, and correctly handle M not divisible by 4.
4) Do NOT add “fake CuTe” no-ops (unused make_shape/make_stride). If CuTe is kept, it must be used for indexing (cute::Tensor / layouts), otherwise remove it.
5) Do not use cuBLAS, do not change precision, do not add PyTorch ops in forward() (must be a single extension call).

G) **Distinguish “correct but slow” from “incorrect semantics”**

For convolution kernels, a kernel may be fully correct yet still be a poor implementation.

Before proposing a correctness fix, classify the kernel into one of these buckets:

1. **Semantic/layout bug**
   - wrong axis order
   - wrong weight layout
   - wrong output-shape formula
   - wrong interpretation of asymmetric or kernel-size-1 dimensions

2. **Fragment/tile bug**
   - wrong CuTe tile mapping
   - wrong partition semantics
   - wrong MMA/GEMM fragment roles
   - wrong smem layout or epilogue masking

Do not misdiagnose a structurally naive kernel as a correctness bug.
If the kernel passes correctness but is still very slow, switch to performance diagnosis.
===============================================================================
PART 4: ASYNC COPY & SYNC
=========================

**Async correctness rules (do these before performance tuning):**
- If you introduce `cp.async` / pipelined stages, every stage must have:
  - a clear “produce” point (copy issued)
  - a clear “consume” point (copy completed + visible)
  - correct barrier/wait placement
- Symptom mapping:
  - stale/previous-tile values → missing wait/barrier
  - fails only when stages > 1 → stage index math or barrier placement bug
- For debugging you may temporarily serialize:
  - stages = 1
  - copy then compute
  - extra sync
  Then restore original pipeline once fixed.

===============================================================================
PART 5: MMA / TiledMMA PITFALLS
===============================

- Atom selection must match arch + dtype.
- Smem layout must match what the MMA path expects (or you must transform).
- Partitioning must align A/B fragments so K is consistent.
- Debug rule: prove A/B fragments contain the intended values *before* blaming MMA.
- CuTe GEMM static-assert failures: debug partition *semantics*, not just types
If compilation fails inside `cute/algorithm/gemm.hpp` with `static_assert` errors involving checks like `size<1>(A) == size<1>(B)`, `size<0>(B) == size<1>(C)`, `size<1>(B) == size<2>(C)`, or similar, treat this as a **fragment-shape mismatch caused by incorrect tiling/partition mapping**. In practice, this usually means one of the `local_tile(...)`, `local_partition(...)`, `Step<...>`, or thread-fragment layouts is assigning the wrong logical role to a dimension. For GEMM, explicitly verify that the fragments presented to `gemm(...)` obey the contract `A = (M_tile, K_tile)`, `B = (K_tile, N_tile)`, and `C/Acc = (M_tile, N_tile)`. Do **not** just tweak types blindly. Instead, inspect whether (1) CTA tiling mapped B through the wrong tiler axis, (2) the shared→compute partition for B accidentally broadcasts or collapses K/N, or (3) the compute micro-tile layout (`tC`, accumulator fragment, or MMA fragment shape) is inconsistent with what CuTe infers. When fixing this class of bug, prefer the smallest structural correction: repair the `Step<...>` mapping, repair the partition direction, or make the compute micro-tile match the implied GEMM fragment shape. After the fix, explicitly state the resulting fragment shapes, e.g. `tCsA = (16,8)`, `tCsB = (8,16)`, `tCrC = (16,16)`, and confirm that they satisfy the GEMM contract before recompiling.
- Implicit-GEMM convolution kernels: verify the logical GEMM mapping end-to-end

For convolution kernels rewritten as implicit GEMM, do NOT stop after making CuTe ranks compile.
A kernel can compile and still be numerically wrong if the logical roles of A, B, and C are inconsistent.

For implicit-GEMM conv2d/conv3d, explicitly write down:

- What GEMM dimensions mean:
  - `M` = output channels per group (or equivalent output feature dimension)
  - `N` = output spatial positions (e.g. `OH*OW`, `OD*OH*OW`)
  - `K` = reduction dimension (e.g. `IC_per_group * KH * KW`, or `IC_per_group * KD * KH * KW`)

- What each operand means:
  - `A(M,K)` = weights / filters
  - `B(K,N)` = im2col-style activation tile
  - `C(M,N)` = output tile

Then prove all three levels are consistent:

1. **Global-memory semantics**
   - Confirm the flattened indexing really matches the intended GEMM mapping.
   - For the activation/im2col operand, verify that decoding `k -> (ic, kh, kw, ...)` and `n -> (oh, ow, ...)` is correct.
   - Check group offsets explicitly.

2. **Shared-memory semantics**
   - State the exact shared-memory tile shapes.
   - For `A`, confirm smem is loaded as `(M_tile, K_tile)`.
   - For `B`, confirm smem is loaded as `(K_tile, N_tile)` — this is a very common bug.
   - Do not accept “same elements, different orientation” as harmless; CuTe GEMM is layout-sensitive.

3. **Thread-fragment semantics**
   - After `local_partition(...)`, explicitly state:
     - `tCsA = (M_frag, K_frag)`
     - `tCsB = (K_frag, N_frag)`
     - `tCrC = (M_frag, N_frag)`
   - If you cannot state these shapes and roles clearly, do not claim the fix is correct.

4. **CTA/output tiling rank sanity**
   - If output `C` is a 2D tensor `(M,N)`, then:
     - CTA tiler must be 2D
     - CTA coord must be 2D
     - `local_tile(...)` / `Step<...>` rank must match that 2D structure
   - Do not pass a 3D coord (e.g. extra `_`) to a 2D tile unless the tensor/tile/projection ranks are explicitly designed for that.

5. **B-operand warning**
   - The most common implicit-GEMM conv bug is fixing `B` by only changing the smem declaration or only changing the store order.
   - That is NOT sufficient unless the downstream `local_partition(...)` and `gemm(...)` now see `B` as `(K_frag, N_frag)`.
   - If `B` was previously treated as `(N,K)` anywhere in the pipeline, re-check:
     - smem shape
     - smem write indices
     - partition direction / `Step<...>`
     - resulting fragment shape seen by `gemm(...)`
6. **Implicit-GEMM conv bug hotspot: B is very often accidentally loaded as (N,K) instead of (K,N)**

For implicit-GEMM convolution, the most common structural bug is:
    - A is treated as (M,K)
    - C is treated as (M,N)
    - but B is accidentally loaded or partitioned as (N,K) somewhere in the pipeline

    This can happen even if:
    - the same values are present
    - shared-memory allocation compiles
    - some partition calls appear shape-compatible

    Required check:
    - state the shared-memory shape of B
    - state the smem write order used to populate B
    - state the fragment shape seen by `gemm(...)`

    Do not claim the bug is fixed unless all three agree that B is logically `(K_tile, N_tile)`.

    If B was changed, re-check:
    - smem declaration
    - smem indexing order
    - local_partition / Step mapping
    - resulting thread fragment role

7. **Do not trust an epilogue-only fix**
   - If max error remains nontrivial after masking stores, the bug is likely in operand mapping or fragment semantics, not just in the ragged-edge store.
   - Edge masking can fix OOB corruption, but it cannot fix a wrong GEMM contraction.

8. **Required proof before declaring success**
   - Provide a short table or statement like:
     - `A tile in smem: (128,8) = (M_tile,K_tile)`
     - `B tile in smem: (8,128) = (K_tile,N_tile)`
     - `C tile in gmem: (128,128) = (M_tile,N_tile)`
     - `tCsA = (8,8)`, `tCsB = (8,8)`, `tCrC = (8,8)` with the correct logical roles
   - Then state why the contraction now matches `C[m,n] += A[m,k] * B[k,n]`.

If a patch fixes compilation by changing ranks, shapes, or tile declarations but does NOT re-prove the A/B/C logical roles above, treat the kernel as still suspect.
===============================================================================
PART 6: MEMORY & OOM PITFALLS (IMPORTANT FOR LLMs)
=================================================

1) The “redundant contiguous() causes OOM” failure mode

Sometimes correctness-check harnesses (especially for non-matmul kernels) accidentally trigger **multiple redundant contiguous conversions** and/or large temporary buffers—causing a crash like:

> Runtime error when checking correctness: **CUDA out of memory. Tried to allocate 6.00 GiB. GPU 0 has a total capacity of 23.49 GiB of which 3.76 GiB is free. Process 392481 has 12.44 GiB memory in use. Including non-PyTorch memory, this process has 6.38 GiB memory in use**

**Guideline for the LLM: treat OOM during correctness-check as a bug in the *test/integration path*, not necessarily in the kernel math.**
Common causes:
- Calling `.contiguous()` on large tensors multiple times in a single check.
- Creating “reference” outputs in multiple dtypes (fp32 + fp16) simultaneously.
- Materializing giant intermediate tensors (e.g., expanded/broadcasted views) instead of using views.
- Copying tensors to GPU repeatedly inside loops (per-test-case allocation churn).

2) Fix OOM WITHOUT changing kernel intent
You may do any of the following (these preserve intent):
- **Cache contiguous buffers**: if you truly need contiguous, do it once and reuse.
- **Avoid duplicate references**: compute reference in-place when possible; free temporaries between runs.
- **Use views instead of materialization**: avoid `.repeat`, `.expand` followed by implicit materialization.
- **Reduce correctness batch size**: validate on smaller sizes first, then scale.
- **Explicitly delete temporaries + synchronize** in the harness between checks when needed.

You must NOT “fix OOM” by rewriting the CuTe kernel into a simpler kernel. The kernel is not the place to “contiguous away” integration problems.

3) A quick diagnostic checklist for this exact error
- Does the harness call `.contiguous()` more than once per input?
- Is there an accidental `.clone()` or `.to(device)` repeated inside a loop?
- Are there multiple large outputs kept alive (not freed) across tests?
- Is the reference path expanding tensors (broadcast) into a full dense materialization?

===============================================================================

END
===
\end{Verbatim}
\end{tcolorbox}

\section{Optimization Prompts}
\label{app:optimization_prompt}

CuTeGen uses workload-specific optimization prompts during the iterative refinement stage. The optimization prompts condition the LLM on the current kernel implementation, the original PyTorch reference code, and runtime measurements from both the PyTorch baseline and the previously generated kernel. Similar to the initial synthesis stage, the prompts require the model to preserve the original functionality and extension structure while implementing exactly one optimization attempt per iteration. This incremental optimization strategy enables each modification to be compiled, validated, and refined independently, preventing numerous debugging stages caused by applying multiple complex optimizations at once.

For Level-1 workloads, the optimization prompt first classifies kernels into categories such as GEMM-like kernels, elementwise kernels, convolution kernels, or other workloads. The prompt then provides workload-specific optimization guidance corresponding to the identified kernel type. For GEMM-like kernels, the prompt emphasizes optimizations such as CTA tiling, shared-memory staging, tensor-core usage, pipelined data movement, and MMA-compatible operand layouts. For convolution kernels, the prompt distinguishes between structurally naive direct implementations and cooperative tiled implementations, encouraging structural rewrites when necessary rather than premature low-level tuning. Elementwise kernels instead prioritize optimizations such as vectorized memory access, fusion of existing operations, grid-stride execution, and occupancy improvements. The prompt also incorporates staged profiling guidance, instructing the model to treat profiling data as secondary evidence and prioritize structural or algorithmic improvements before profiler-driven parameter tuning.

For Level-2 workloads, the optimization prompt shifts focus from optimizing individual operators to optimizing the computation graph as a whole. The prompt explicitly instructs the model to identify producer and consumer operations, intermediate tensors, and opportunities for algebraic simplification or kernel fusion. In particular, the prompt prioritizes eliminating unnecessary intermediate tensors, reducing kernel launches, fusing epilogues, and simplifying shape-trivial reductions when mathematically valid. This design encourages the model to optimize end-to-end memory movement and execution structure rather than applying isolated operator-level optimizations.

Across both prompts, several hard constraints are enforced to preserve semantic correctness. The prompts prohibit changes in numerical precision, prohibit fallback to PyTorch operators or vendor libraries such as cuBLAS or cuDNN, and require the generated implementation to preserve the original output semantics and tensor shapes. Together, these prompts operationalize the workload-aware optimization strategy described in Section~\ref{sec:opt} while maintaining stable iterative refinement behavior.
\begin{tcolorbox}[
  colback=white,
  colframe=black,
  boxrule=0.8pt,
  breakable,
  left=4pt,
  right=4pt,
  top=4pt,
  bottom=4pt,
  title={CuTe Primitive Operation Optimization Prompt}
]
\begin{Verbatim}[breaklines=true,fontsize=\footnotesize]
You're given the following CUTE/CUDA source code:
```
<NODE_PRV_SRC>
```

with the following PyTorch reference code
```
<NODE_REF_SRC>
```
PYTORCH_REFERENCE_TIME: <REFERENCE_TIME>
PREVIOUS_SOURCE_TIME: <PREVIOUS_SOURCE_TIME>

Your task: Optimize this code using CUTE/CUDA.
IMPORTANT: Implement ONLY ONE optimization attempt. Pick ONE optimization that is not already implemented.
You must output REAL compilable code (not pseudocode) and preserve the original code structure:
- Keep the same function names, arguments, and return types.
- The optimized output architecture is named ModelNew with custom CUTE/CUDA kernel(s).
- Output ONLY the new CUTE/CUDA code + torch glue code. No extra text, no explanations, no testing code.
- Output the entire new code in ONE CODEBLOCK (wrap in ``` and ```).

Before optimizing, infer what type of kernel this is:
(A) Matrix Multiply / GEMM-like kernel (e.g., C = A @ B, A*C + A*D, batched GEMM, etc.)
(B) Activation / Elementwise kernel (e.g., ReLU, tanh, sigmoid, GELU, add+relu, bias+relu, clamp, etc.)
(C) Convolution kernel (e.g., 1D/2D/3D convolution, depthwise convolution, grouped convolution, pointwise convolution, spatial convolution, conv-like sliding-window operators)
(D) Other (if none of the above fits, apply safe general GPU optimizations only)

Your optimization choice MUST match the kernel type.
======================
WHEN PROFILER DATA IS AVAILABLE

Use profiler data as secondary evidence, not as the main optimization driver.

Choose optimizations in this order:
1) kernel-specific optimization principles
2) missing structural optimizations
3) strong profiler evidence
4) minor launch or tuning adjustments

Profiler data should help you:
- confirm bottlenecks
- break ties between plausible optimizations
- avoid obviously bad choices
- validate whether the current kernel structure is weak

Use profiler data strongly only when it gives a clear signal, such as:
- very high registers/thread causing low achieved occupancy
- very high runtime despite reasonable occupancy
- very low shared-memory use in a reuse-heavy kernel
- clear memory-bandwidth limitation
- obvious launch configuration problems
- explicit profiler warnings directly related to the next optimization

For GEMM and convolution kernels, structural and algorithmic decisions are usually more important than raw counters unless we already have plausible optimizations implemented.
If the kernel is still structurally weak, prefer fixing the structure over tuning around profiler metrics.


If runtime remains high, ask whether the kernel is still fundamentally direct or structurally inefficient.

===========================================================
If the kernel is (A) Matrix Multiply / GEMM-like:
Pick ONE of the following effective optimizations (only one):
1) Threadblock-level tiling (CTA tiling for M/N/K)
2) Warp-level tiling
3) Thread-level tiling / vectorized loads
4) Tensor Core mma operations (only if data types + layout allow it)
5) Shared-memory bank conflict reduction (swizzling / layout transforms)
6) Pipelining / multi-stage loads (cp.async if supported)
7) Split-K (only if K dimension is very large, make sure it is implemented)
8) Asynchronous loads/stores
9) For GEMM-like kernels where one dimension is much smaller than the others, tile over the large output dimensions, keep the small dimension inside the CTA (unroll if small), and avoid square-GEMM or split-K strategies but make sure you implement this optimization.
HIGH-PRIORITY GEMM OPTIMIZATION: Shared-memory operand layout for MMA/WMMA

For GEMM-like kernels, treat shared-memory layout as a high-priority structural optimization, especially for B.

When staging operands into shared memory:
- Prefer layouts that match the MMA/WMMA load layout.
- For row-major A[M,K] and row-major B[K,N], a strong default is:
  - smA[M_tile, K_tile]
  - smB[K_tile, N_tile]
- Prefer contiguous/vectorized copies from global memory into shared memory, e.g. float4 or cp.async 16-byte copies.
- Avoid manually scattering B into a transposed shared-memory layout unless the MMA instruction explicitly benefits from that layout.
- If using WMMA, choose the shared-memory layout and the wmma::row_major / wmma::col_major tag consistently.
- For row-major B[K,N], prefer storing B in shared memory as K_tile x N_tile and loading it as wmma::row_major when valid.
- Add shared-memory padding/skew when needed to reduce bank conflicts, e.g. SMEM_B_STRIDE = N_TILE + SKEW.
- This optimization should be considered before minor tuning changes such as changing block size, changing K_TILE, or changing launch parameters.

Rules for GEMM:
- Pay close attention to matrix operand dimensions (M, N, K) and layouts/strides and how they compare to each other to deliver specific optimizations.
- Pick the strongest optimization that matches the shapes.
- If the input matrices have a specific characteristic, use them in a way that benefits the performance. For example, if a matrix is symmetric the number of reads could be reduced.
- There might be information about the matrices in comments or the reference code (such as if a matrix is symmetric etc) you can optimize only for those cases to make kernels fast but DO NOT make assumptions yourself without having provable and explicit information in the reference.
- Do NOT change precision.
- Do NOT use cuBLAS/cuDNN or other pre-optimized libraries.
- Do NOT fall back to PyTorch operators for any case.
- It is acceptable to specialize to the exact shapes listed.

===========================================================
If the kernel is (B) Activation / Elementwise:
Matmul optimizations like tensor cores, split-K, warp tiling for M/N/K, and matrix swizzling are NOT applicable.
Pick ONE of the following elementwise optimizations (only one):
1) Vectorized global memory loads/stores (float4/half2/etc.) when contiguous + aligned
2) Grid-stride loop to cover arbitrary numel efficiently
3) Reduce kernel launch overhead: fuse simple elementwise ops already present in the code (ONLY if they already exist in the original kernel computation; do not invent new semantics)
4) Use fast math intrinsics safely (only for tanh/sigmoid/GELU-like kernels; keep outputs correct/stable)
5) Improve tail handling + bounds checks to avoid invalid memory accesses
6) Minimize register pressure / unnecessary temporaries
7) Improve occupancy by choosing a better block size (128/256/512) WITHOUT changing semantics

Rules for elementwise kernels:
- Treat the tensor as a 1D array of length L = x.numel().
- Memory access must be coalesced.
- Avoid using shared memory unless there is clear data reuse (elementwise usually has none).
- Keep correct behavior for all elements, including tail elements when L is not divisible by vector width.
- Do NOT use or fall back to PyTorch ops even for special cases.
- Do NOT change precision.
- It is acceptable to specialize to the exact dtype and contiguous layout if the original code already assumes it.

===========================================================
If the kernel is (C) Convolution kernel:

First decide whether the current kernel is still structurally direct.

Treat it as structurally direct if any of the following are true:
- one thread computes one output, or only a very small number of outputs
- accumulation is still dominated by nested scalar loops over input channels and kernel elements
- neighboring outputs repeatedly reload reusable input activations from global memory
- shared memory is used only partially and does not enable strong cooperative reuse
- CuTe is used mainly for indexing/layout convenience rather than true tiled partitioning
- the kernel is still far from reference performance despite local optimizations

If the kernel is still structurally direct:
DO NOT apply a micro-optimization.
Instead, perform exactly ONE structural rewrite that moves the kernel toward a cooperative tiled implementation.

Valid structural rewrites:
1) CTA-level output/spatial tiling
2) shared-memory staging of reusable input tiles
3) cooperative thread computation across output tiles
4) real CuTe tiled tensor partitioning
5) implicit GEMM restructuring, if the convolution shape supports it

If the kernel is already cooperative and tiled:
Implement exactly ONE optimization from the list below:
1) improve output/spatial tiling
2) improve shared-memory reuse
3) improve cooperative thread mapping
4) add register blocking
5) improve memory coalescing
6) add pipelined staged loads
7) add tensor-core MMA only if layout, dtype, and shape clearly support it

Shape-aware rules:
- Choose the optimization based on the actual convolution shape, not generic conv advice.
- When input channels are very small, avoid over-investing in channel-reduction tiling.
- When output volume is very large, prioritize output/spatial tiling and input reuse.
- When the kernel remains direct after an earlier attempt, prefer structural rewrite or implicit GEMM over further local refinement.
- Do not apply fancy load pipelining or buffering unless the kernel already has a strong tiled structure.

Hard rules:
- Preserve exact convolution semantics
- Preserve stride / padding / dilation / groups behavior
- Do NOT change precision
- Do NOT use cuBLAS or cuDNN
- Do NOT fall back to PyTorch operators
- Implement exactly ONE optimization attempt
- Prefer structural improvement over cosmetic/local optimization
===========================================================
If the kernel is (D) Other / unclear:
Apply ONE safe general GPU optimization:
- vectorized load/store OR
- better bounds handling OR
- grid-stride loop
Do NOT introduce GEMM-specific assumptions.

===========================================================
Additional hard constraints:
- Keep the given source code structure the same, including function names, arguments, return types, and extension loading style.
- Generate complete working code that compiles.
- Do not output multiple versions; implement exactly one optimization attempt.
- Do not add any benchmark or test code.
- Do not add extra commentary outside the code block.
- Pay attention to alignment, indexing correctness, and avoiding out-of-bounds accesses.

Make sure the optimized code implements the same functionality as the previous version.

===========================================================
Performance red flags that indicate a naive kernel

When profiler data is available, treat the following as strong signals that the kernel
is structurally weak:

- very high duration relative to the reference
- high registers/thread causing occupancy collapse
- full or high occupancy but still huge runtime
- little or no shared-memory usage in a kernel with heavy reuse potential
- compute-heavy direct accumulation with no tensor-core/GEMM structure
- huge grid with one-thread-per-output or tiny register blocking only

Interpretation guide:
- High occupancy does NOT mean a kernel is good.
- CuTe usage does NOT mean the kernel is tiled.
- Shared-memory staging of one operand alone does NOT guarantee speedup.
- If runtime is still very high, ask whether the kernel is still fundamentally direct.

\end{Verbatim}
\end{tcolorbox}

\begin{tcolorbox}[
  colback=white,
  colframe=black,
  boxrule=0.8pt,
  breakable,
  left=4pt,
  right=4pt,
  top=4pt,
  bottom=4pt,
  title={CuTe Multi-Operation Optimization Prompt}
]
\begin{Verbatim}[breaklines=true,fontsize=\footnotesize]
You're given the following CUTE/CUDA source code:
```
<NODE_PRV_SRC>
```

with the following PyTorch reference code
```
<NODE_REF_SRC>
```
PYTORCH_REFERENCE_TIME: <REFERENCE_TIME>
PREVIOUS_SOURCE_TIME: <PREVIOUS_SOURCE_TIME>

Your task: Optimize this code using CUTE/CUDA.
IMPORTANT: Implement ONLY ONE optimization attempt. Pick ONE optimization that is not already implemented.
You must output REAL compilable code (not pseudocode) and preserve the original code structure:
- Keep the same function names, arguments, and return types.
- The optimized output architecture is named ModelNew with custom CUTE/CUDA kernel(s).
- Output ONLY the new CUTE/CUDA code + torch glue code. No extra text, no explanations, no testing code.
- Output the entire new code in ONE CODEBLOCK (wrap in ``` and ```).
===========================================================
OPTIMIZATION PRINCIPLE
===========================================================

This is a multi-operation workload. Do NOT optimize it as only a GEMM, only a convolution, or only an elementwise kernel.

Before choosing the CUDA optimization, infer the full forward computation as a graph:

1. Identify the producer operation:
   - linear / matmul / batched matmul
   - convolution / depthwise convolution / grouped convolution
   - normalization
   - reduction
   - elementwise chain
   - other

2. Identify all consumer operations after the producer:
   - bias/add/scale
   - activation
   - clamp
   - pooling
   - sum/mean/max/min
   - logsumexp
   - reshape/view/transpose
   - normalization
   - other reductions

3. Identify intermediate tensors:
   - Which tensors are written to global memory?
   - Which tensors are read only once?
   - Which tensors can be eliminated?

Prefer optimizations that reduce total memory traffic and kernel launches.

===========================================================
FIRST PRIORITY: ALGEBRAIC SIMPLIFICATION
===========================================================

Before low-level CUDA tuning, check if the forward computation can be simplified mathematically.

You SHOULD apply algebraic simplification when it is exactly equivalent.

Examples of valid simplifications:
- reduction over a dimension of size 1 is identity
- max over a singleton dimension is identity
- mean over a singleton dimension is identity
- logsumexp over a singleton dimension is identity
- sum(linear(x), dim=output_dim) can become:
      x @ sum(weight, dim=0) + sum(bias)
- consecutive elementwise operations can be computed in registers
- bias/add/scale/activation can be fused into the producer epilogue
- if a temporary tensor is written once and read once, try to eliminate it

Do NOT change semantics.
Do NOT change output shape.
Do NOT change dtype/precision.
Do NOT change broadcasting behavior.
Do NOT assume commutativity or associativity unless the transformation is mathematically exact for the given operation.

If an operation is shape-trivial, such as max/mean/logsumexp over a size-1 dimension, simplify it.

===========================================================
SECOND PRIORITY: FUSION
===========================================================

Prefer real fusion over fake fusion.

Bad pattern:
    kernel1 writes tmp to global memory
    kernel2 reads tmp from global memory
    kernel2 writes output

Good pattern:
    compute producer value in registers
    apply following operations immediately
    write final output once

Valid fusion attempts:
1. Fuse bias/add/scale/activation into producer writeback.
2. Fuse consecutive elementwise operations into one kernel.
3. Fuse reduction epilogue when the reduction is small, local, or shape-trivial.
4. Remove one intermediate tensor that is written once and read once.
5. Collapse a producer + reduction chain when mathematically possible.
6. Cache/precompute parameter-only reductions if valid for the timed forward path.

===========================================================
PARAMETER-ONLY REDUCTION RULE
===========================================================

If the computation repeatedly reduces only model parameters, such as:
    sum(weight, dim=...)
    sum(bias)
and those reduced values are reused for every batch element, do not recompute them per output element.

Good options:
- compute the parameter reduction once in a separate kernel, then use it in the main kernel
- cache reduced parameter tensors as buffers if the model weights are fixed during timed inference
- fuse the reduced parameter use into the main computation

Avoid recomputing a full weight reduction inside every output computation.

===========================================================
KERNEL TYPE GUIDANCE
===========================================================

After checking algebraic simplification and fusion, choose the relevant low-level strategy.

(A) GEMM / Linear / Matmul-like:
- Prefer epilogue fusion first.
- If the output is immediately reduced, consider whether the matmul can be replaced by a smaller equivalent computation.
- If full GEMM is still needed, use CTA tiling, shared memory, register blocking, and coalesced memory access.
- Tensor cores are allowed only if dtype/layout/precision make them valid.
- Do NOT use cuBLAS, cuDNN, or PyTorch operators.

(B) Elementwise chains:
- Treat tensors as flat contiguous arrays when valid.
- Use grid-stride loops.
- Use vectorized loads/stores when alignment and tail handling are correct.
- Compute all elementwise operations in registers.
- Store only the final result.

(C) Reductions:
- If reducing over a singleton dimension, remove the reduction.
- If reducing over a small dimension, use block-level or warp-level reduction.
- If reducing over a large dimension, use cooperative parallel reduction.
- Do not serialize a large reduction in one thread.
- Avoid writing large intermediate tensors just to reduce them later.

(D) Convolution / Conv-like:
- First preserve exact convolution semantics:
  stride, padding, dilation, groups, axis order, weight layout, and output shape.
- If followed by bias/activation/scale, fuse into writeback.
- If followed by elementwise operations, compute them before storing.
- If followed by reductions, check whether the reduction can be fused or algebraically simplified.
- If the current kernel is direct and slow, prefer a structural improvement:
  output tiling, shared-memory input reuse, cooperative computation, or implicit GEMM if shape supports it.
- Do NOT use cuDNN or PyTorch fallback.

(E) Normalization / Softmax / Pooling:
- Do not blindly fuse complex reductions if unsafe.
- But if the operation is over a singleton dimension, shape-trivial, or already local to each output element, simplify or fuse it.
- Preserve numerical behavior and output shape.

===========================================================
PROFILER GUIDANCE
===========================================================

Use profiler data as secondary evidence.

Optimization priority:
1. mathematical simplification / fusion opportunity
2. eliminating intermediates and kernel launches
3. structural CUDA/CuTe improvement
4. profiler-guided tuning
5. minor block-size/register tweaks

High occupancy does NOT mean the kernel is good.
CuTe usage does NOT mean the kernel is truly tiled.
Shared memory usage does NOT automatically mean there is useful reuse.
If runtime is still high, ask whether the code is still writing/reading unnecessary intermediates.

===========================================================
HARD CONSTRAINTS
===========================================================

- Do NOT change the computation.
- Do NOT remove operations unless they are mathematically identity/trivial for the given shapes.
- Do NOT change operation order unless mathematically identical.
- Do NOT change precision.
- Do NOT use PyTorch ops inside forward().
- Do NOT use cuBLAS/cuDNN/CUTLASS device operators.
- Do NOT generate fallback paths using PyTorch.
- Do NOT add testing code.
- Do NOT output explanations outside the codeblock.
- Make sure the code compiles.
- Make sure the code returns the same shape as the reference.

Make sure the optimized code implements the same functionality as the previous version.
\end{Verbatim}
\end{tcolorbox}

\section{Comparison of CuTe and CUDA Representations for Square GEMM}
\label{app:cute_vs_cuda}
To better understand the role of the target representation in LLM-based kernel 
generation, we compare two minimal square matrix multiplication kernels produced 
under similar prompting conditions: one in standard CUDA and one in CuTe. The goal 
is not to claim that the initial CuTe kernel is already highly optimized, but to 
show that it places the model in a more favorable region of the design space from 
the outset.

As shown in Figure \ref{fig:simp_cuda}, the simplest raw CUDA kernel follows a thread-per-output-element pattern: each 
thread computes a single output entry using direct global-memory indexing and a 
full inner-product loop over the reduction dimension. While easy to generate, this 
form encodes little of the structure required for high-performance GEMM. It lacks 
explicit notions of tiling, warp-level decomposition, shared-memory staging, and 
separation between data movement and computation. As a result, moving from this 
baseline to an efficient implementation requires introducing several tightly 
coupled transformations---blocking, mapping to thread hierarchies, shared-memory 
staging, synchronization, and tensor-core usage---effectively restructuring the 
kernel rather than refining it incrementally.

\vspace{10pt}
\noindent
\begin{lstlisting}[language=C++,basicstyle=\ttfamily\footnotesize,frame=single,
aboveskip=3pt,belowskip=3pt]
__global__ void matmul_kernel(const float* __restrict__ A,
                              const float* __restrict__ B,
                              float* __restrict__ C,
                              int M, int K, int N) {
    int row = blockIdx.y * blockDim.y + threadIdx.y;  // [0, M)
    int col = blockIdx.x * blockDim.x + threadIdx.x;  // [0, N)

    if (row < M && col < N) {
        float sum = 0.0f;
        for (int k = 0; k < K; ++k) {
            sum += A[row * K + k] * B[k * N + col];
        }
        C[row * N + col] = sum;
    }
}
\end{lstlisting}
\captionsetup{width=\linewidth,font=footnotesize}
\refstepcounter{figure}\label{fig:simp_cuda}
{Figure \thefigure: }{Na\"{\i}ve CUDA implementation of square matrix multiplication. Each thread computes a single output element using direct global-memory accesses and a full reduction over $K$. While functionally correct and easy to generate, this form lacks explicit structure for tiling, shared-memory reuse, or hierarchical work decomposition, making it a weak starting point for performance optimization.}
\label{fig:simp_cuda}
\vspace{10pt}

In contrast, the simplest CuTe kernel (Figure \ref{fig:simple_cute}) already exposes the core structural axes of 
GEMM. It explicitly represents the problem shape $(M,N,K)$, constructs tensor views, 
defines a CTA tiler, and partitions computation across blocks and threads, with 
shared-memory buffers and thread layouts present from the outset. Although not yet 
optimized, this formulation begins from a tiled and partitioned structure, encouraging 
the model to reason directly about blocking, decomposition, and data movement. Even at this early stage, this structural advantage translates into measurable 
performance differences: the initial CuTe kernel achieves a $4.5\times$ speedup over 
the naïve CUDA baseline (approximately 0.45$\times$ vs.\ 0.1$\times$ relative to the PyTorch 
reference), despite neither implementation being fully optimized.

Crucially, CuTe separates logical tensor structure from its physical mapping onto 
threads and memory. This reduces reliance on manual index arithmetic and allows 
modifications to tiling, layouts, and thread mappings without rewriting the entire 
kernel. In contrast, similar changes in raw CUDA typically require substantial 
reworking of indexing logic and control flow, increasing the likelihood of 
correctness regressions.

Overall, enforcing CuTe generation is not merely a syntactic choice. For square GEMM, 
it changes the form of the initial solution: the raw CUDA kernel is correct but 
structurally minimal, whereas the CuTe kernel already reflects the abstractions used 
in high-performance implementations. This provides a stronger and more stable 
starting point for subsequent optimization.
\vspace{10pt}
\noindent
\begin{lstlisting}[language=C++,basicstyle=\ttfamily\footnotesize,frame=single,
aboveskip=3pt,belowskip=3pt,breaklines=true,columns=fullflexible]
template <class TA, class TB, class TC>
__global__ void
gemm_nn_device(cute::Shape<int,int,int> shape_MNK,
               TA const* A, cute::Stride<int,int> dA,
               TB const* B, cute::Stride<int,int> dBt,  // Transposed view stride for B: (N,K) with stride (1,N)
               TC      * C, cute::Stride<int,int> dC)
{
  using namespace cute;

  // Unpack shape
  int const M = get<0>(shape_MNK);
  int const N = get<1>(shape_MNK);
  int const K = get<2>(shape_MNK);

  // Global memory tensors
  // A is (M,K) row-major
  Tensor mA = make_tensor(make_gmem_ptr(A), make_shape(M, K), dA);
  // B is given as (K,N) row-major, create a transposed view (N,K) with stride (1, N)
  Tensor mB = make_tensor(make_gmem_ptr(B), make_shape(N, K), dBt);
  // C is (M,N) row-major
  Tensor mC = make_tensor(make_gmem_ptr(C), make_shape(M, N), dC);

  // Tile shapes: (Mtile, Ntile, Ktile)
  auto cta_tiler = make_shape(Int<128>{}, Int<128>{}, Int<8>{});
  auto cta_coord = make_coord(blockIdx.x, blockIdx.y, _);

  // Tile tensors from GMEM
  Tensor gA = local_tile(mA, cta_tiler, cta_coord, Step<_1, X, _1>{});   // (128, 8) along M and K
  Tensor gB = local_tile(mB, cta_tiler, cta_coord, Step<X, _1, _1>{});   // (128, 8) along N and K (B is (N,K))
  Tensor gC = local_tile(mC, cta_tiler, cta_coord, Step<_1, _1, X>{});   // (128,128) along M and N

  // Shared memory tiles
  __shared__ TA smemA[128*8];
  __shared__ TB smemB[128*8];
  Tensor sA = make_tensor(make_smem_ptr(smemA), make_shape(Int<128>{}, Int<8>{}));
  Tensor sB = make_tensor(make_smem_ptr(smemB), make_shape(Int<128>{}, Int<8>{}));

  // Thread layouts
  auto tA = make_layout(make_shape(Int<32>{}, Int<8>{}));
  auto tB = make_layout(make_shape(Int<32>{}, Int<8>{}));
  auto tC = make_layout(make_shape(Int<16>{}, Int<16>{}));

  // Partition GMEM tiles for each thread to load to SMEM
  Tensor tAgA = local_partition(gA, tA, threadIdx.x);
  Tensor tAsA = local_partition(sA, tA, threadIdx.x);
  Tensor tBgB = local_partition(gB, tB, threadIdx.x);
  Tensor tBsB = local_partition(sB, tB, threadIdx.x);

  // Partition SMEM tiles for compute and GMEM C tile for store
  Tensor tCsA = local_partition(sA, tC, threadIdx.x, Step<_1, X>{});
  Tensor tCsB = local_partition(sB, tC, threadIdx.x, Step<X, _1>{});
  Tensor tCgC = local_partition(gC, tC, threadIdx.x, Step<_1, _1>{});

  // Accumulator
  Tensor tCrC = make_tensor_like(tCgC);
  clear(tCrC);

  // K loop over tiles
  int const K_TILE_MAX = size<2>(tAgA);
  for (int k_tile = 0; k_tile < K_TILE_MAX; ++k_tile) {
    // Load A and B tiles from GMEM to SMEM
    copy(tAgA(_,_,k_tile), tAsA);
    copy(tBgB(_,_,k_tile), tBsB);
    __syncthreads();

    // Compute GEMM on the tiles
    gemm(tCsA, tCsB, tCrC);
    __syncthreads();
  }

  // Write back to GMEM: C = 1.0 * Acc + 0.0 * C
  axpby(1.0f, tCrC, 0.0f, tCgC);
}
\end{lstlisting}

\captionsetup{type=figure,width=\linewidth,font=footnotesize}
\phantomsection
\captionof{figure}{Initial CuTe implementation of square matrix multiplication. Unlike the naïve CUDA baseline, this kernel is already expressed in terms of GEMM-relevant abstractions, including explicit problem shapes, CTA tiling, tensor views, shared-memory tiles, and thread-level partitions. Although still simple and not fully optimized, this representation gives the model a stronger starting point by making blocking, data movement, and hierarchical work decomposition explicit from the outset.}
\label{fig:simple_cute}
\section{Case Study: Square Matrix Multiplication}
\label{sec:case_study_square_gemm}

We illustrate CuTeGen's synthesis behavior through a representative case study: \emph{square} matrix multiplication, where the input matrices are square and of equal dimension (i.e., $A \in \mathbb{R}^{N \times N}$ and $B \in \mathbb{R}^{N \times N}$). Square GEMM is a canonical GPU workload and serves as a useful reference point because it exposes the full complexity of high-performance matrix multiplication—including hierarchical tiling, shared-memory staging, asynchronous memory movement, and tensor-core utilization—while avoiding confounding effects from highly irregular shapes. The example below highlights how CuTeGen produces an implementation that matches established GEMM design patterns and how CuTe's structured tensor abstractions help guide the model toward correct indexing, consistent tiling, and stable iterative refinement.

\paragraph{Hierarchical tiling and work decomposition.}
The generated kernel adopts a multi-level tiling strategy. At the threadblock level, each CUDA thread block (256 threads, organized as eight warps) computes a fixed output tile of $128 \times 128$ elements and iterates over the reduction dimension in $K$-tiles of size $16$, as shown in Figure~\ref{lst:tiling}. Within each block, warps are arranged as a $2 \times 4$ grid, where each warp computes a $64 \times 32$ output region. Finally, each warp decomposes its region into $16 \times 16$ tensor-core micro-tiles, consistent with the WMMA fragment configuration illustrated in Figure~\ref{lst:wmma_compute}. The tile sizes, warp decomposition, and tensor layouts are internally consistent across shared-memory staging, WMMA fragment dimensions, and global-memory indexing, ensuring correct dimensional alignment and stable execution. This hierarchy (threadblock tile $\rightarrow$ warp tile $\rightarrow$ tensor-core tile) mirrors standard high-performance GEMM implementations and is generated automatically within the CuTeGen refinement loop.
\vspace{10pt}
\noindent
\begin{lstlisting}[language=C++,basicstyle=\ttfamily\footnotesize,frame=single,
aboveskip=3pt,belowskip=3pt]
using namespace cute;
// Threadblock (CTA) tile: (Mtile, Ntile, Ktile)
auto cta_tiler = make_shape(Int<128>{}, Int<128>{}, Int<16>{});
auto cta_coord = make_coord(blockIdx.x, blockIdx.y, _);

// Global tensors: A(M,K), B(N,K) as a transposed view, C(M,N)
Tensor mA = make_tensor(make_gmem_ptr(A), make_shape(M, K), dA);
Tensor mB = make_tensor(make_gmem_ptr(B), make_shape(N, K), dBt);
Tensor mC = make_tensor(make_gmem_ptr(C), make_shape(M, N), dC);

// Tiles corresponding to this block
Tensor gA = local_tile(mA, cta_tiler, cta_coord, Step<_1, X, _1>{});
Tensor gB = local_tile(mB, cta_tiler, cta_coord, Step<X, _1, _1>{});
Tensor gC = local_tile(mC, cta_tiler, cta_coord, Step<_1, _1, X>{});
\end{lstlisting}

\captionof{figure}{Threadblock tiling and CuTe tensor views (square GEMM). The generated kernel assigns each block a $128\times128$ output tile and iterates over $K$ in tiles of $16$. CuTe shape/stride tensor views make dimensional intent explicit and help stabilize iterative refinement.}
\label{lst:tiling}
\vspace{10pt}

\paragraph{Double-buffered shared-memory staging.}
To reduce global-memory latency and improve overlap between memory movement and compute, the generated kernel allocates two shared-memory stages for both $A$ and $B$ tiles, as illustrated in Figure~\ref{lst:smem_double_buffer}. While tensor-core computation proceeds on one stage, the next $K$-tile is prefetched into the alternate stage. The shared-memory layouts incorporate padding (``skew'') along leading dimensions to mitigate bank conflicts and preserve $16$-byte alignment for vectorized memory transactions. Through iterative debugging and optimization, the model converged on a configuration in which shared-memory strides, tile dimensions, and WMMA fragment sizes remain mutually consistent  (Figure~\ref{lst:wmma_compute}), avoiding the indexing and layout mismatches commonly observed in naïve CUDA generation. Although the kernel uses TF32 WMMA tensor-core instructions internally, we do not classify this as a mixed-precision violation during manual auditing because the kernel operates on FP32 tensors and preserves FP32 accumulation and output semantics. This reflects NVIDIA’s standard accelerated execution path for FP32 GEMM workloads on on Ampere- and Ada-class GPUs, where WMMA/MMA tensor-core operations use TF32 internally for FP32 inputs. In contrast, kernels that explicitly lower operand or storage precision to FP16/BF16 are treated as reduced-precision implementations and excluded from reported speedup numbers. The resulting structure reflects non-trivial performance engineering decisions typically associated with hand-tuned kernels.

\vspace{10pt}
\noindent
\begin{lstlisting}[language=C++,basicstyle=\ttfamily\footnotesize,frame=single,
aboveskip=2pt,belowskip=2pt]
constexpr int K_TILE = 16;
constexpr int SKEW_A = 8, SKEW_B = 8;
constexpr int SMEM_A_STRIDE = K_TILE + SKEW_A;   // padded leading dimension
constexpr int N_TILE = 128;
constexpr int SMEM_B_STRIDE = N_TILE + SKEW_B;   // padded leading dimension

extern __shared__ __align__(16) unsigned char smem[];

// Double-buffered shared memory: A0,B0,A1,B1
float* smA0 = reinterpret_cast<float*>(smem);
float* smB0 = reinterpret_cast<float*>(smA0 + 128 * SMEM_A_STRIDE);
float* smA1 = reinterpret_cast<float*>(smB0 + K_TILE * SMEM_B_STRIDE);
float* smB1 = reinterpret_cast<float*>(smA1 + 128 * SMEM_A_STRIDE);

// CuTe shared-memory tensor views with explicit strides (skew/padding)
Tensor sA0 = make_tensor(make_smem_ptr(smA0),
                         make_shape(Int<128>{}, Int<K_TILE>{}),
                         make_stride(Int<SMEM_A_STRIDE>{}, Int<1>{}));
Tensor sB0 = make_tensor(make_smem_ptr(smB0),
                         make_shape(Int<K_TILE>{}, Int<128>{}),
                         make_stride(Int<SMEM_B_STRIDE>{}, Int<1>{}));
\end{lstlisting}
\captionof{figure}{Double-buffered shared-memory staging with skewed strides. Two stages are allocated for both $A$ and $B$ tiles to enable pipelined prefetching. Padding (skew) reduces bank conflicts and helps preserve $16$-byte alignment for vectorized transactions (e.g., \texttt{cp.async} of 16 bytes).}
\label{lst:smem_double_buffer}
\vspace{10pt}

\paragraph{Inline PTX and asynchronous global-to-shared copies.}
A particularly notable aspect of the generated implementation is its correct emission of architecture-specific inline PTX instructions to implement asynchronous global-to-shared memory transfers. As shown in Figure~\ref{lst:cp_async_helpers}, the kernel defines device-side wrappers around the SM80+ \texttt{cp.async} instruction, issuing $16$-byte transactions together with explicit commit and wait-group synchronization. These instructions integrate directly with the double-buffered staging mechanism described above, enabling pipelined data movement across $K$-tiles. Importantly, the implementation includes architecture guards via \texttt{\_\_CUDA\_ARCH\_\_} checks, falling back to a synchronous vectorized copy (e.g., \texttt{float4}) on pre-SM80 GPUs. This demonstrates that the agent is capable of synthesizing low-level assembly-backed primitives correctly while preserving portability across hardware generations.

\vspace{10pt}
\noindent
\begin{lstlisting}[language=C++,basicstyle=\ttfamily\footnotesize,frame=single,
aboveskip=3pt,belowskip=3pt]
namespace wmma = nvcuda::wmma;
constexpr int WMMA_M = 16, WMMA_N = 16, WMMA_K = 8;

// Accumulators cover a 64x32 warp tile (4x2 WMMA tiles)
wmma::fragment<wmma::accumulator, WMMA_M, WMMA_N, WMMA_K, float> c_frag[4][2];
for (int i = 0; i < 4; ++i)
  for (int j = 0; j < 2; ++j)
    wmma::fill_fragment(c_frag[i][j], 0.0f);

// Compute on the current K tile in steps of WMMA_K
for (int kk = 0; kk < K_TILE; kk += WMMA_K) {
  for (int i = 0; i < 4; ++i) {
    int a_row = warp_m_base + i * WMMA_M;
    wmma::fragment<wmma::matrix_a, WMMA_M, WMMA_N, WMMA_K,
                   wmma::precision::tf32, wmma::row_major> a_frag;
    wmma::load_matrix_sync(a_frag, baseA_smem + a_row * SMEM_A_STRIDE + kk, SMEM_A_STRIDE);

    for (int j = 0; j < 2; ++j) {
      int b_col = warp_n_base + j * WMMA_N;
      wmma::fragment<wmma::matrix_b, WMMA_M, WMMA_N, WMMA_K,
                     wmma::precision::tf32, wmma::row_major> b_frag;
      wmma::load_matrix_sync(b_frag, baseB_smem + kk * SMEM_B_STRIDE + b_col, SMEM_B_STRIDE);

      wmma::mma_sync(c_frag[i][j], a_frag, b_frag, c_frag[i][j]);
    }
  }
}
\end{lstlisting}
\captionof{figure}{Tensor-core compute loop using WMMA. Each warp computes a $64\times32$ region using $16\times16\times8$ WMMA operations. The loop iterates over the $K$-tile in two steps of $8$ and accumulates results in FP32 registers before storing to global memory. The implementation uses NVIDIA’s standard TF32 WMMA execution path for FP32 GEMM workloads on Ada and Ampere GPUs.}
\label{lst:wmma_compute}
\vspace{10pt}

\paragraph{Early CuTe-only kernel generation.}
Before converging to the optimized implementation described above, the initial kernel generated by CuTeGen relied exclusively on CuTe abstractions without explicit tensor-core WMMA primitives, inline PTX instructions, or asynchronous memory staging. An excerpt from this early version is shown in Figure~\ref{lst:cute}. The kernel expresses the GEMM computation entirely through CuTe tensor views, tiling constructs, and high-level operations such as \texttt{copy}, \texttt{gemm}, and \texttt{axpby}. 

Although this version achieved only moderate performance (approximately $0.45\times$ the reference PyTorch implementation), it was functionally correct and exhibited consistent dimensional reasoning across global-memory layouts, shared-memory tiling, and thread-level partitioning. This CuTe-based formulation provided a stable baseline from which subsequent optimization iterations introduced tensor-core WMMA instructions, asynchronous memory transfers, and double-buffered shared-memory staging, ultimately producing the optimized kernel discussed earlier with a speedup of approximately $1.16\times$.

\vspace{10pt}
\noindent
\begin{lstlisting}[language=C++,basicstyle=\ttfamily\footnotesize,frame=single,
aboveskip=2pt,belowskip=2pt]
#if __CUDA_ARCH__ >= 800
__device__ __forceinline__ void cp_async_16(void* smem_ptr, const void* gmem_ptr) {
  unsigned smem_addr = static_cast<unsigned>(__cvta_generic_to_shared(smem_ptr));
  asm volatile("cp.async.cg.shared.global [%0], [%1], 16;" :: "r"(smem_addr), "l"(gmem_ptr));
}
__device__ __forceinline__ void cp_async_commit() {
  asm volatile("cp.async.commit_group;");
}
__device__ __forceinline__ void cp_async_wait_all() {
  asm volatile("cp.async.wait_group 0;");
}
#else
__device__ __forceinline__ void cp_async_16(void* smem_ptr, const void* gmem_ptr) {
  *reinterpret_cast<float4*>(smem_ptr) = *reinterpret_cast<const float4*>(gmem_ptr);
}
__device__ __forceinline__ void cp_async_commit() {}
__device__ __forceinline__ void cp_async_wait_all() {}
#endif
\end{lstlisting}
\captionsetup{type=figure,font=footnotesize,width=\linewidth}
\captionof{figure}{\textbf{Architecture-gated asynchronous copy helpers.} For SM80+, the kernel uses inline PTX \texttt{cp.async} to copy $16$ bytes per instruction from global to shared memory and synchronizes via commit/wait group operations. For pre-SM80 targets, it falls back to a synchronous 16-byte vectorized copy and treats commit/wait as no-ops.}
\label{lst:cp_async_helpers}
\vspace{10pt}

\paragraph{Role of CuTe abstractions in guiding correct generation.}
More broadly, CuTe functions as a structured intermediate representation that constrains the synthesis space while still allowing progressively lower-level optimizations. Although the final implementation incorporates WMMA tensor-core primitives and inline PTX asynchronous copy instructions, its overall organization continues to follow the CuTe tensor-view and tiling abstractions illustrated in Figure~\ref{lst:tiling}. These abstractions make dimensional relationships explicit—including consistent interpretation of $M$, $N$, and $K$ and block-to-tile mappings—reducing ambiguity in indexing and memory layout transformations while supporting incremental optimization without requiring a complete restructuring of the kernel.

\vspace{20pt}
\noindent
\begin{lstlisting}[language=C++,basicstyle=\ttfamily\footnotesize,frame=single,
aboveskip=2pt,belowskip=2pt]
  // Shared memory tiles
  __shared__ TA smemA[128*8];
  __shared__ TB smemB[128*8];
  Tensor sA = make_tensor(make_smem_ptr(smemA), make_shape(Int<128>{}, Int<8>{}));
  Tensor sB = make_tensor(make_smem_ptr(smemB), make_shape(Int<128>{}, Int<8>{}));

  // Thread layouts
  auto tA = make_layout(make_shape(Int<32>{}, Int<8>{}));
  auto tB = make_layout(make_shape(Int<32>{}, Int<8>{}));
  auto tC = make_layout(make_shape(Int<16>{}, Int<16>{}));

  // Partition GMEM tiles for each thread to load to SMEM
  Tensor tAgA = local_partition(gA, tA, threadIdx.x);
  Tensor tAsA = local_partition(sA, tA, threadIdx.x);
  Tensor tBgB = local_partition(gB, tB, threadIdx.x);
  Tensor tBsB = local_partition(sB, tB, threadIdx.x);

  // Partition SMEM tiles for compute and GMEM C tile for store
  Tensor tCsA = local_partition(sA, tC, threadIdx.x, Step<_1, X>{});
  Tensor tCsB = local_partition(sB, tC, threadIdx.x, Step<X, _1>{});
  Tensor tCgC = local_partition(gC, tC, threadIdx.x, Step<_1, _1>{});

  // Accumulator
  Tensor tCrC = make_tensor_like(tCgC);
  clear(tCrC);

  // K loop over tiles
  int const K_TILE_MAX = size<2>(tAgA);
  for (int k_tile = 0; k_tile < K_TILE_MAX; ++k_tile) {
    // Load A and B tiles from GMEM to SMEM
    copy(tAgA(_,_,k_tile), tAsA);
    copy(tBgB(_,_,k_tile), tBsB);
    __syncthreads();

    // Compute GEMM on the tiles
    gemm(tCsA, tCsB, tCrC);
    __syncthreads();
  }
  axpby(1.0f, tCrC, 0.0f, tCgC);
}
\end{lstlisting}
\captionsetup{type=figure,font=footnotesize,width=\linewidth}
\phantomsection
\captionof{figure}{\textbf{Initial CuTe-only GEMM kernel generated by CuTeGen.}
This early version relies entirely on CuTe tensor abstractions without explicit tensor-core WMMA primitives or asynchronous memory instructions. While functionally correct, it achieves lower performance than the optimized kernel but provides a stable basis for subsequent refinement.}
\label{lst:cute}

\section{Additional Experimental Results}
\label{ab_app}
\paragraph{Per-Level Comparison with CudaForge.}

Tables~\ref{tab:cutegen_vs_cudaforge_l1} and
\ref{tab:cutegen_vs_cudaforge_l2}
provide a more detailed breakdown of the comparison between CuTeGen and
CudaForge across KernelBench Level-1 and Level-2 workloads.
These results further illustrate the different optimization regimes of the two
benchmark levels discussed in Section~\ref{sec:exp}.

On Level-1 workloads, where PyTorch dispatches directly to highly optimized
vendor libraries such as cuBLAS and cuDNN, CuTeGen substantially outperforms
CudaForge across all reported metrics.
CuTeGen achieves a 1.32$\times$ average speedup compared to 0.61$\times$ for
CudaForge and produces the best-performing kernel in 49\% of the tasks versus
only 2\% for CudaForge.
CuTeGen also achieves substantially higher correctness and Fast1 rates,
suggesting that the structured CuTe representation and debugging guidelines helps stabilize iterative
optimization on structurally complex kernels.

On Level-2 workloads, both systems benefit from operator fusion opportunities,
and the gap between the two systems narrows.
CuTeGen still achieves higher average speedup overall (2.12$\times$ versus
1.19$\times$), but CudaForge attains a higher Fast1 percentage.
This behavior is consistent with the discussion in section \ref{sec:exp}: once kernel
fusion is achieved, the remaining optimization headroom from low-level tuning
is often smaller than in Level-1 workloads.
\begin{table}[t]
\centering
\caption{
Comparison of CuTeGen and CudaForge on KernelBench Level-1 workloads.
We report average speedup, precision lowering rate (\%), correctness (\%), Fast1 (\%), speedup on Fast1 tasks, percentage of best-performing kernels (\%).
}
\label{tab:cutegen_vs_cudaforge_l1}

\footnotesize

\begin{tabular*}{\linewidth}{@{\extracolsep{\fill}}lccccccc}
\toprule
\textbf{Tool}
& \textbf{Speedup} $\uparrow$
& \textbf{Low Precision} $\downarrow$
& \textbf{Correct} $\uparrow$
& \textbf{Fast1} $\uparrow$
& \textbf{Speedup On Fast1} $\uparrow$
& \textbf{Best} $\uparrow$\\
\midrule
CudaForge 
& 0.61$\times$& 5\% & 66\%
& 16\% & 1.23$\times$ & 2\%\\

\midrule
\textbf{CuTeGen}
& 1.32$\times$ & 0\% & 94\%
& 50\% & 1.65$\times$ & 49\%\\

\bottomrule
\end{tabular*}
\end{table}

\begin{table}[t]
\centering
\caption{
Comparison of CuTeGen and CudaForge on KernelBench Level-2 workloads.
We report average speedup, precision lowering rate (\%), correctness (\%), Fast1 (\%), speedup on Fast1 tasks, percentage of best-performing kernels (\%).
}
\label{tab:cutegen_vs_cudaforge_l2}

\footnotesize

\begin{tabular*}{\linewidth}{@{\extracolsep{\fill}}lccccccc}
\toprule
\textbf{Tool}
& \textbf{Speedup} $\uparrow$
& \textbf{Low Precision} $\downarrow$
& \textbf{Correct} $\uparrow$
& \textbf{Fast1} $\uparrow$
& \textbf{Speedup On Fast1} $\uparrow$
& \textbf{Best} $\uparrow$\\
\midrule
CudaForge 
& 1.19$\times$& 33\% & 59\%
& 32\% & 1.67$\times$ & 27\%\\

\midrule
\textbf{CuTeGen}
& 2.12$\times$ & 0\% & 76\%
& 22\% & 2.71$\times$ & 21\%\\

\bottomrule
\end{tabular*}
\end{table}

\if 0
\begin{table}[t]
\centering

\footnotesize

\begin{tabular*}{\linewidth}{@{\extracolsep{\fill}}lccccccc}
\toprule
\textbf{Method}
& \multicolumn{3}{c}{\textbf{Level 1}}
& \multicolumn{3}{c}{\textbf{Level 2}}
& \textbf{Cost (USD)} $\downarrow$ \\
\cmidrule(lr){2-4} \cmidrule(lr){5-7}
& Speedup $\uparrow$
& Fast1 $\uparrow$
& Correct. $\uparrow$
& Speedup $\uparrow$
& Fast1 $\uparrow$
& Correct. $\uparrow$
& \\
\midrule

CudaForge 
& 0.72$\times$& 18\% & 69\%
& 1.41$\times$ & 47\% & 72\%
& 1.41 \\

\midrule
\textbf{CuTeGen}
& 1.35$\times$ & 50\% & 94\%
& 2.12$\times$ & 22\% & 76\%
& 1.69 \\

\bottomrule
\end{tabular*}
\end{table}

\fi

\paragraph{Additional Ablation Study Results}

Tables~\ref{tab:main-results-fast1} and
\ref{tab:main-results-correctness}
provide additional ablation results for the CuTeGen variants discussed in
Section~\ref{sec:exp}.
While Table~\ref{tab:ablation} reports average speedup across categories, the
tables in this section additionally report Fast1 rates and correctness percentages.

The results further support the conclusions of the main ablation study.
Operating in CuTe consistently improves both correctness and Fast1 rates across
most workload categories, particularly for convolution and
reduction/normalization kernels where structural tiling and memory-layout
reasoning are important.
Similarly, delaying profiling integration generally produces stronger overall
results than introducing profiling feedback from the outset, especially on
structurally complex workloads such as matrix multiplication and convolution.

\begin{table}[htbp]
\centering
\caption{
Fast1 over PyTorch on KernelBench Level-1 tasks for different CuTeGen variants, grouped by kernel category.
}
\label{tab:main-results-fast1}
\resizebox{\linewidth}{!}{
\begin{tabular}{lcccccc}
\toprule
\textbf{Method}
& \textbf{MatMul}
& \textbf{Activation Function}
& \textbf{Convolution}
& \textbf{Reduction/Normalization}
& \textbf{Loss}
& \textbf{Total} \\
\midrule

CuTeGen profiling from start
& 16\%
& \textbf{100\%}
& 18\%
& 50\%
& \textbf{100\%}
& 41\%
\\
CuTeGen without profiling
& 21\%
& 93\%
& 15\%
& 66\%
& \textbf{100\%}
& 43\%

\\
CuTeGen with cuda
& \textbf{26\%}
& \textbf{100\%}
& 18\%
& 53\%
& \textbf{100\%}
& 42\%

\\

\textbf{CuTeGen}
& \textbf{26\%}
& \textbf{100\%}
& \textbf{21\%}
& \textbf{77\%}
& \textbf{100\%}
& \textbf{50\%}

\\
\bottomrule
\end{tabular}
}
\end{table}

\begin{table}[htbp]
\centering
\caption{
Correctness percentage over PyTorch on KernelBench Level-1 tasks for different CuTeGen variants, grouped by kernel category.
}
\label{tab:main-results-correctness}
\resizebox{\linewidth}{!}{
\begin{tabular}{lcccccc}
\toprule
\textbf{Method}
& \textbf{MatMul}
& \textbf{Activation Function}
& \textbf{Convolution}
& \textbf{Reduction/Normalization}
& \textbf{Loss}
& \textbf{Total} \\
\midrule

CuTeGen profiling from start
& 84\%
& \textbf{100\%}
& 89\%
& 73\%
& \textbf{100\%}
& 85\%
\\
CuTeGen without profiling
& 89\%
& \textbf{100\%}
& 89\%
& \textbf{85\%}
& \textbf{100\%}
& 91\%

\\
CuTeGen with cuda
& \textbf{100\%}
& \textbf{100\%}
& 85\%
& 73\%
& \textbf{100\%}
& 88\%

\\

\textbf{CuTeGen}
& \textbf{100\%}
& \textbf{100\%}
& \textbf{92\%}
& \textbf{85\%}
& \textbf{100\%}
& \textbf{94\%}

\\
\bottomrule
\end{tabular}
}
\end{table}

\section{Licenses of External Assets}

Our experiments use several publicly available software assets and benchmarks.
PyTorch is distributed under a BSD-style (BSD-3-Clause) license.
CUTLASS and its CuTe abstraction layer are distributed under the BSD-3-Clause license.
Nsight Compute is used under NVIDIA’s developer tools license agreement.
KernelBench and CudaForge are used according to the licenses and terms provided in their public repositories.
Kernel generation and refinement are performed using the GPT-5 API subject to OpenAI’s applicable terms of use.
We cite all external assets used in our experiments and use them in accordance with their publicly available licenses and terms of use.

\end{document}